\PassOptionsToPackage{table, prologue}{xcolor}

\pdfoutput=1

\documentclass[11pt]{article}

\usepackage[]{emnlp2021}

\usepackage{times}
\usepackage{latexsym}

\usepackage[T1]{fontenc}

\usepackage[utf8]{inputenc}

\usepackage{microtype}

\usepackage{graphicx}
\usepackage{caption}
\usepackage{subcaption}

\definecolor{lightgray}{gray}{0.9}

\usepackage{xspace}
\usepackage{booktabs}
\usepackage{multicol}
\usepackage{amsmath}
\usepackage{multirow}
\usepackage{graphicx}
\usepackage{enumitem}
\usepackage{mdwlist}
\usepackage{textcomp}
\usepackage{siunitx}
\usepackage{longtable}
\usepackage{pbox}
\usepackage{amssymb}
\usepackage{xspace}
\usepackage{movie15}

\newcommand{\fref}[1]{Figure~\ref{#1}}

\title{Exploring Linguistic Style Matching in Online Communities: The Role of Social Context and Conversation Dynamics}

\author{Aparna Ananthasubramaniam\thanks{~~denotes equal contribution}, Hong Chen\footnotemark[1], Jason Yan\footnotemark[1], Kenan Alkiek\footnotemark[1], Jiaxin Pei\footnotemark[1], \\
\bf{Agrima Seth\footnotemark[1], Lavinia Dunagan\footnotemark[1], Minje Choi\footnotemark[1], Benjamin Litterer\footnotemark[1] \and David Jurgens} \\
        School of Information, University of Michigan \\ \texttt{\{akananth,hongcc,jasonyan,kalkiek,pedropei,}\\
        \texttt{agrima,laviniad,minje,blitt,jurgens\}@umich.edu}}

\begin{document}
\maketitle

\begin{abstract}
 Linguistic style matching (LSM) in conversations can be reflective of several aspects of social influence such as power or persuasion. However, how LSM relates to the outcomes of online communication on platforms such as Reddit is an unknown question.
In this study, we analyze a large corpus of two-party conversation threads in Reddit where we identify all occurrences of LSM using two types of style: the use of function words and formality. Using this framework, we examine how levels of LSM differ in conversations depending on several social factors within Reddit: post and subreddit features, conversation depth, user tenure, and the controversiality of a comment. Finally, we measure the change of LSM following loss of status after community banning.
Our findings reveal the interplay of LSM in Reddit conversations with several community metrics, suggesting the importance of understanding conversation engagement when understanding community dynamics.


\end{abstract}
\section{Introduction}

Social influence can be subtle. When two persons converse, their interpersonal dynamics can lead to one person adopting the language of the other. For example, in settings where one person has higher status or power, the lower-status person may unconsciously begin mirroring the language of the other~\citep{danescu2012echoes}. This process has been described as \textit{accommodation} \cite{giles2007communication} or \textit{linguistic style matching (LSM)} \cite{niederhoffer2002linguistic} and can reflect the underlying influence that individuals have on each other \cite{chartrand1999chameleon}. Past work has primarily focused on how linguistic influence changes relative to the identities of the speakers. However, the larger social context in which a conversation happens also plays a role in determining whether an individual may be influential. Here, we perform a large-scale study of linguistic influence to test how specific types of social context influence the level of accommodation.

Past work in the social sciences has studied accommodation to understand the influence and social power dynamics in specific settings, like job interviews (applicants and interviewers) \cite{willemyns1997accent} and academic context (students and faculty)\cite{jones1999strategies}. Also, LSM has been studied to understand group dynamics~\citep{gonzales2010language} and negotiations~\citep{ireland2014negotiations}. Work in NLP has operationalized these theories to test accommodation theory in new domains. Typically, these works adopt some tests for measuring influence in language and have shown these measures correlate with known social differences. However, it is yet unknown how LSM occurs in conversations in online community platforms and differs by community dynamics.

Our work examines the larger context in which linguistic influence occurs. Using a large sample of 2.3 million conversations from Reddit and two measures of linguistic influence, we test how the level of linguistic influence correlates with conversational outcomes, such as conversation length and even the continued presence of a person in a community. Further, we examine how specific social and contextual factors influence the rates of linguistic influence. For instance, we discover that the controversy level of the parent comment can lead to different dynamics of style matching in the conversation threads.

This paper offers the following three contributions. First, we systematically compare complementary measures of accommodation, showing clear evidence of style accommodation in Reddit conversations. Second, we draw the relationships of several social factors that affect LSM, including levels of engagement, the popularity of the content, and tenure within a subreddit. 
Third, we demonstrate the use of LSM to measure the loss of status through the banning of subreddits. We have released all code and data for full reproducibility.\footnote{\url{https://github.com/davidjurgens/style-influence}}


\section{Accommodation and its Measurement}
In this section, we discuss communication accommodation theory and associated sociolinguistic research to outline the accommodation of communicative behavior based on perceived social power dynamics. Subsequently, we explore the concept of linguistic style matching and methods adopted by researchers to quantify this phenomenon. We also investigate various factors that contribute to LSM variations and their strategic uses.

\subsection{Accommodation Theory as Social Influence}

When two individuals engage in social interaction, they may either converge or diverge in their communicative behavior. The Communication Accommodation Theory (CAT) suggests that the degree of convergence or divergence is affected by the relative social power between the interlocutors \cite{xu2018not}. Asymmetric convergence is more likely to occur in situations where there is a power imbalance between the interlocutors. Individuals with lower social power or status are more likely to adapt their communication style to align with those in higher or dominant positions \cite{muir2016characterizing}. For instance, Puerto Ricans in New York City during the 1970s, who were perceived to hold less power than African Americans, adopted the dialect of African Americans to converge with their more powerful counterparts \cite{wolfram1974sociolinguistic}.

Social power has been often found to be an important determinant of degrees of accommodation \cite{giles19911, ng1993power} and interactants of differential social power or social status can act in a complementary fashion \cite{street1991accommodation}.

\subsection{Linguistic Style Matching}
Linguistic alignment is a pervasive phenomenon that occurs in human communication where interactants unconsciously coordinate their language usage. This coordination, described as convergence in the psycholinguistic theory of communication accommodation, involves aspects such as word choice, syntax, utterance length, pitch, and gestures \cite{giles19911}. Linguistic style matching (LSM) is a specific manifestation of linguistic alignment, wherein individuals unconsciously match their speaking or writing styles during conversations \cite{ireland2011language}. Unlike content accommodation, LSM focuses on stylistic accommodation, examining how things are communicated rather than what they communicate.


Individuals strategically negotiate their language style to decrease social distance, seek approval, and accommodate each other. LSM can also reflect the level of common understanding and conceptualization of the conversation topic between speakers. The degree of LSM can indicate social power dynamics as indicated by \cite{giles2007communication}. Empirical evidence from recent studies \cite{danescu2012echoes} showed that participants with less power (such as lawyers or non-administrative roles in Wikipedia) exhibit greater coordination in conversational behavior than participants with high power (such as justices or administrators). Additionally, \citet{noble2015centre} identified a positive correlation between linguistic accommodation and social network centrality, which effect can be greater than the effect of power status distinction. Studies by \citet{muir2016characterizing, muir2017linguistic} further show that individuals in a lower position of power tend to accommodate their linguistic style to match that of their higher-power counterparts during face-to-face communication as well as computer-mediated communication.

The variance in LSM can be attributed to various social and psychological factors and can be triggered for different purposes. Linguistic alignment may signal likability and agreement, relate to seeking approval or arise from social desirability. Higher levels of accommodation in social behaviors are found to be associated with increased feelings of affiliation, liking, and successful interpersonal relationships \cite{bayram2019diplomatic}. 
Thus, linguistic alignment can be strategically employed to establish relationship initiation and stability \cite{ireland2011language}, increase group cohesion, and task performance \cite{gonzales2010language}, and assist in negotiations \cite{taylor2008linguistic}. Furthermore, alignment has been found to enhance persuasiveness, motivating listeners to adopt healthier practices \cite{cialdini2001science} while in some cases like presidential debates, it has been perceived as more aggressive \cite{romero2015mimicry}. The degree of matching may differ based on context and individual factors.




\section{Data}

Reddit is a popular social media platform with a forum-based interface. It allows users to interact with dispersed individuals who share similar experiences or topics of interest. Our dataset to study LSM spans from July 2019 to December 2022 and includes 35M users and 500K subreddits.

Using the Pushshift Reddit Dataset which contains the full history of comments aggregated on a monthly basis~\citep{baumgartner2020pushshift}, we construct conversation threads from the comments and filter those that satisfy the following conditions: (1) the conversation chain consists of exactly two users; (2) the beginning of the conversation chain must be a root comment which does not have a parent comment; and (3) the lengths of a conversation chain must between 3 and 100.
These conditions allow us to capture conversation dynamics between exactly two users without any interference. Our resulting dataset contains 16,893,013 conversation turns (or comments) across 2,305,775 conversation chains from 68,788 subreddits.

\section{How should we measure linguistic influence?}
\label{sec:measure}

Computational work has proposed multiple approaches for both what to measure and how to measure linguistic influence. In this section, we aim to build intuition for what the two measures of accommodation---using function words and formality---are operationalizing.

\begin{figure}[t!]
\centering
    \begin{subfigure}[b]{0.2\textwidth}
        \centering
        \includegraphics[width=\textwidth]{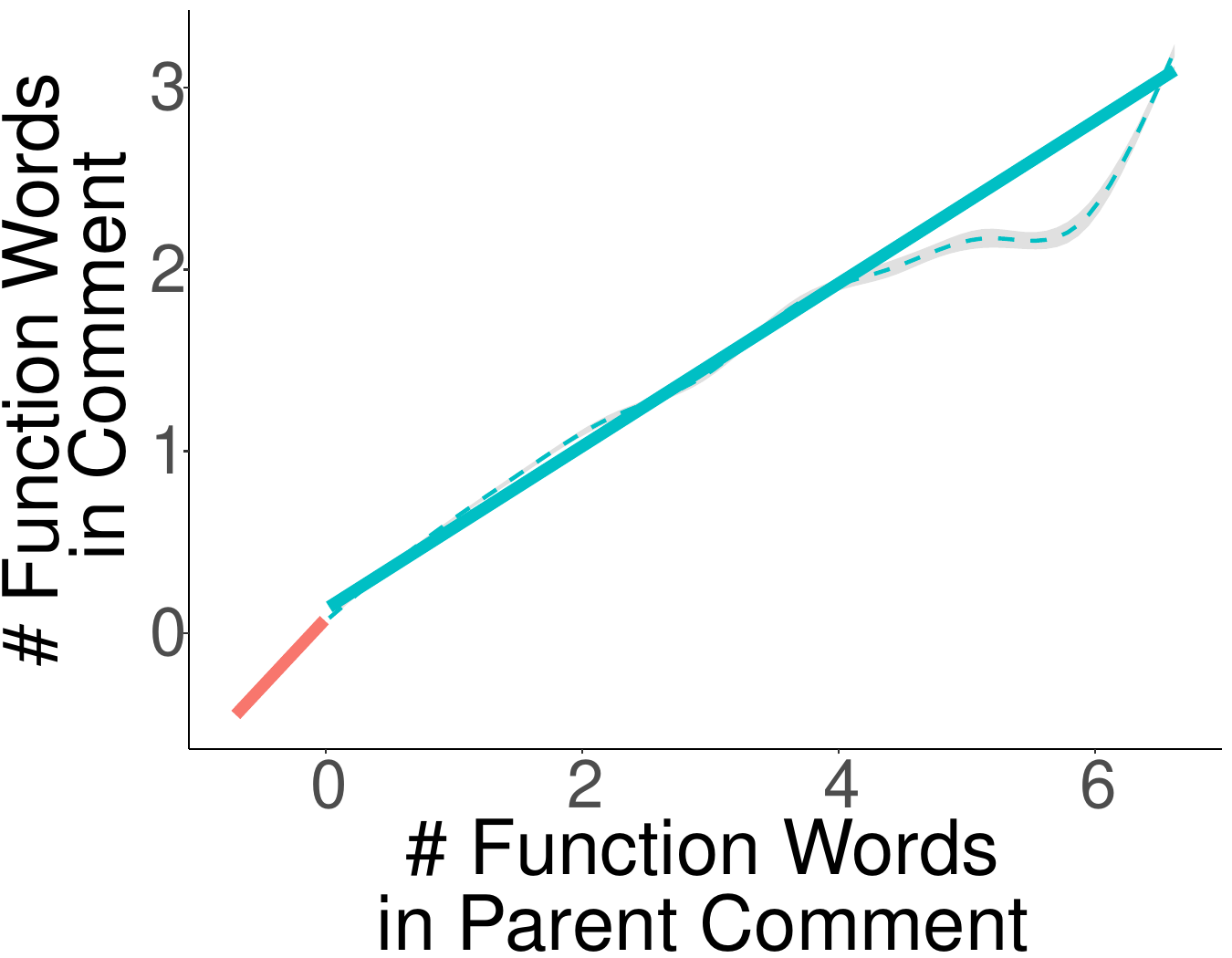}
        \caption{Accommodation, \# Function Words}
        \label{function-lsm}
    \end{subfigure}
    \hfill
    \begin{subfigure}[b]{0.2\textwidth}
        \centering
        \includegraphics[width=\textwidth]{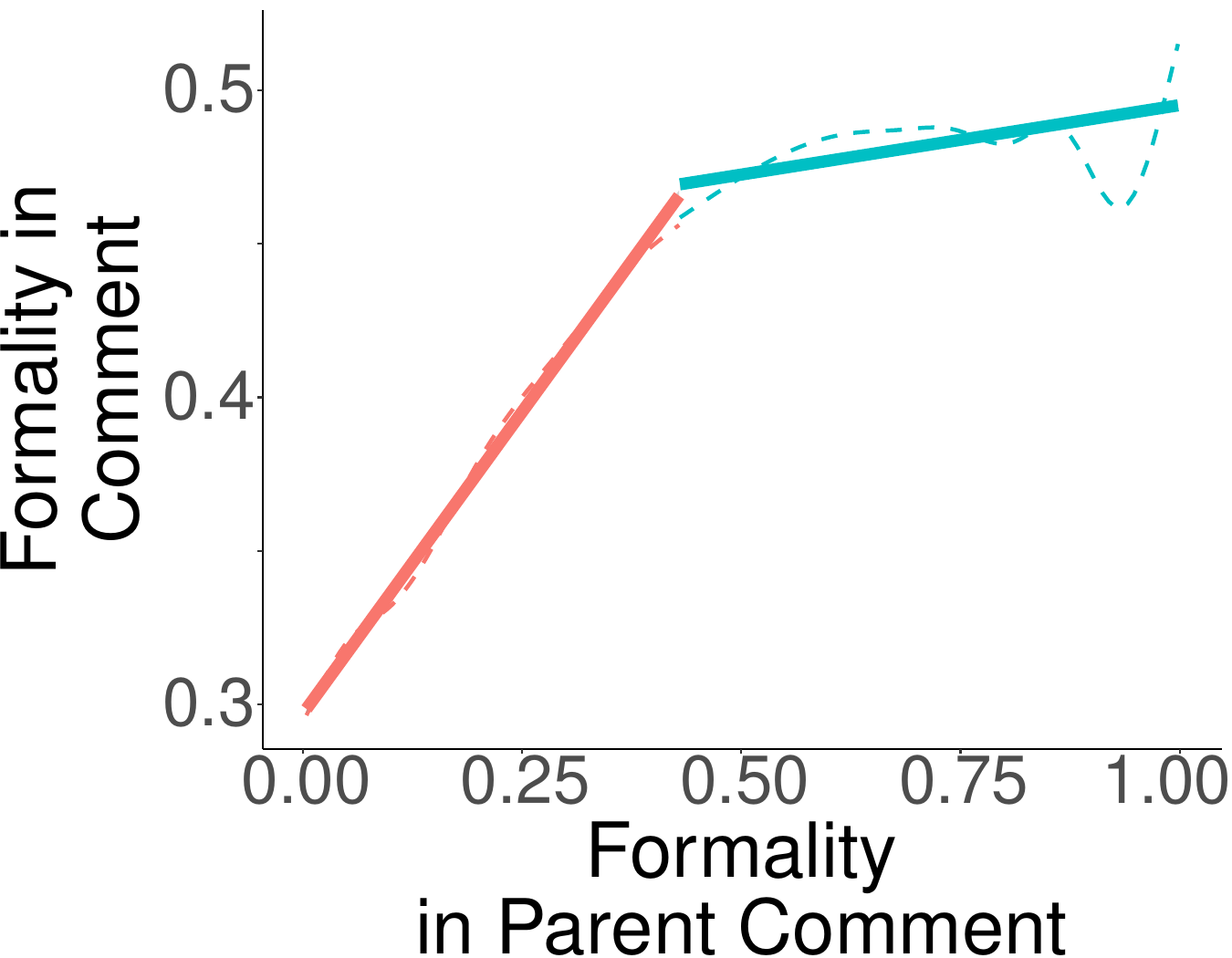}
        \caption{Accommodation, Formality}
        \label{formality-lsm}
    \end{subfigure}
    \caption{Commenters on Reddit accommodate to the (a) \# function words and b) formality of the comment they are replying to. Typically, the level of accommodation is higher when responding to posts with below-average (red) than above-average (blue) style.}
    \label{fig:markers}
\end{figure}

\subsection{Linguistic Style Markers}

Our study  measures linguistic influence with two complementary style markers. We use the notation $m$ to refer to a marker throughout.

\paragraph{Marker 1: Function Words} 
Function words (e.g. pronouns, prepositions, articles, and auxiliary words) are primarily employed unconsciously and frequently and incorporate social knowledge for comprehension and usage \cite{meyer1999representations, ireland2010language}. 
Prior computational studies of linguistic accommodation have measured linguistic influence by tracking the relative frequencies of function words across conversation turns \citep{danescu2011mark, babcock2014latent, gonzales2010language}.
Function words reflect \textit{how} content is expressed, rather than what specific content is expressed (e.g., content words) and are thought to be a better proxy for unconscious language processing \cite{tausczik2010psychological}. Here, we use the function words defined by the Linguistic Inquiry and Word Count (LIWC) lexicon \cite{pennebaker2001linguistic,pennebaker2007expressive}. 

\paragraph{Marker 2: Formality} 
Individuals adopt a specific register that is appropriate to their position in the social context, real or desired \cite{niederhoffer2002linguistic}. A commonly varied register is the level of formality used when speaking to another. The level of formality shown by a speaker is known to reflect the speaker's opinion towards a topic or their closeness to the listener \citep{hovy1979language}. 
Unlike function words, variation in formality often requires conscious processing to select the appropriate phrasing in a given circumstance. As a result, it offers a complementary view into how a speaker influences another through shifting the conversation towards a more formal or informal register.

Here, we measure formality using a supervised classification model. The model is a fine-tuned RoBERTa-based classifier \citep{liu2019roberta} trained on the GYAFC~\citep{rao2018gyafc} and Online Formality Corpus~\citep{pavlick2016formality} datasets; we use the model available from the Hugging Face API\footnote{\url{https://huggingface.co/s-nlp/roberta-base-formality-ranker}}. Both datasets contain social media text and the reported model performance is high for both blogs and Q\&A text (Spearman's $\rho$>0.7).
Using this classifier, each comment's formality is measured on a continuous scale in [0,1].

Importantly, these style variables are related; function word frequency also changes in more formal contexts, where articles and prepositions typically become more common while pronouns and interjections become less common \citep{Heylighen1999FormalityOL}. 
Content word-based measures of style and function word counts are thought to capture the same latent style variables, i.e., they are interchangeable at a stylometric level \citep{Grieve_2023_47_77}. 

\subsection{Measuring Linguistic Influence}

At a high-level, linguistic influence (also referred to as LSM or accommodation in this paper) is measured by testing whether the value for some measure $m$ of a comment made by user $a$ is predictive of the value of $m$ in the reply to that comment by user $b$. 
%
%
Therefore, one straightforward way to measure accommodation is with linear regression: $m_b \sim \beta_0 + \beta_1 m_a$ where $\beta_0$ reflects the baseline level of the measure (e.g., the average formality) and $\beta_1$ measures the level of accommodation (e.g., the average increase in formality associated with a 1-unit increase in the formality of the parent comment). 
However, as \citet{xu2018not} note, the characteristics of a comment are likely influenced by other unrelated factors such as the length of the comment or the number of turns in the conversation. Indeed, they show that unless one controls for such factors, linguistic influence may be overestimated. 
Therefore, we used a mixed-effects regression to control for comment $a$ and $b$'s length in tokens (fixed effects $L_a$, $L_b$), the number of replies $r_{b \rightarrow a}$ that $b$ has made to $a$ so far in the conversation.
To capture individual and community-level variation, we include  random effects to control for the effect of the subreddit $s$; these random effects let us control for differences in the norms of communities (e.g., some communities are more/less formal) to test for relative changes in $m$. 
Linguistic accommodation is modeled as 
\begin{align*}
m_b \sim & \beta_0 + \beta_1 m_a + \\
& \beta_2 L_a + \beta_3 L_b + \beta_4 r_{b \rightarrow a} + (1 | s)
\end{align*}
where $\beta_1$ measures the level of accommodation.






\subsection{Results} 


We first observe clear evidence of accommodation in both style markers: parent comments with more function words receive replies with more function words (Figure~\ref{function-lsm}), and more formal parent comments receive more formal replies (Figure~\ref{formality-lsm}). 
For comments where we have the text of the original post, we observe accommodation even after controlling for the author and original post's style markers, suggesting that users may accommodate to the style of the person they are interacting with in the comment thread. However, this effect plateaus when the parent comment has above-average levels of a style marker, suggesting a potential threshold for the impact of parent comment style on reply style. This attenuation of effect may be the result of several mechanisms, including regression to the mean or an author modulating their replies according to their own personal style (i.e., a more extreme parent comment may trigger greater modulation).

Second, the two style markers are almost perfectly uncorrelated, suggesting that they measure distinct constructs. In order to calculate the correlation between these two measures, we randomly sample 1,000 subsets of the conversation turns and calculate the extent of accommodation in function words and formality in that subset.
The correlation between the function-word- and formality-based accommodation scores is -0.00171. 

\begin{figure}[t]
\centering
    \begin{subfigure}[b]{0.23\textwidth}
        \centering
        \includegraphics[width=\textwidth]{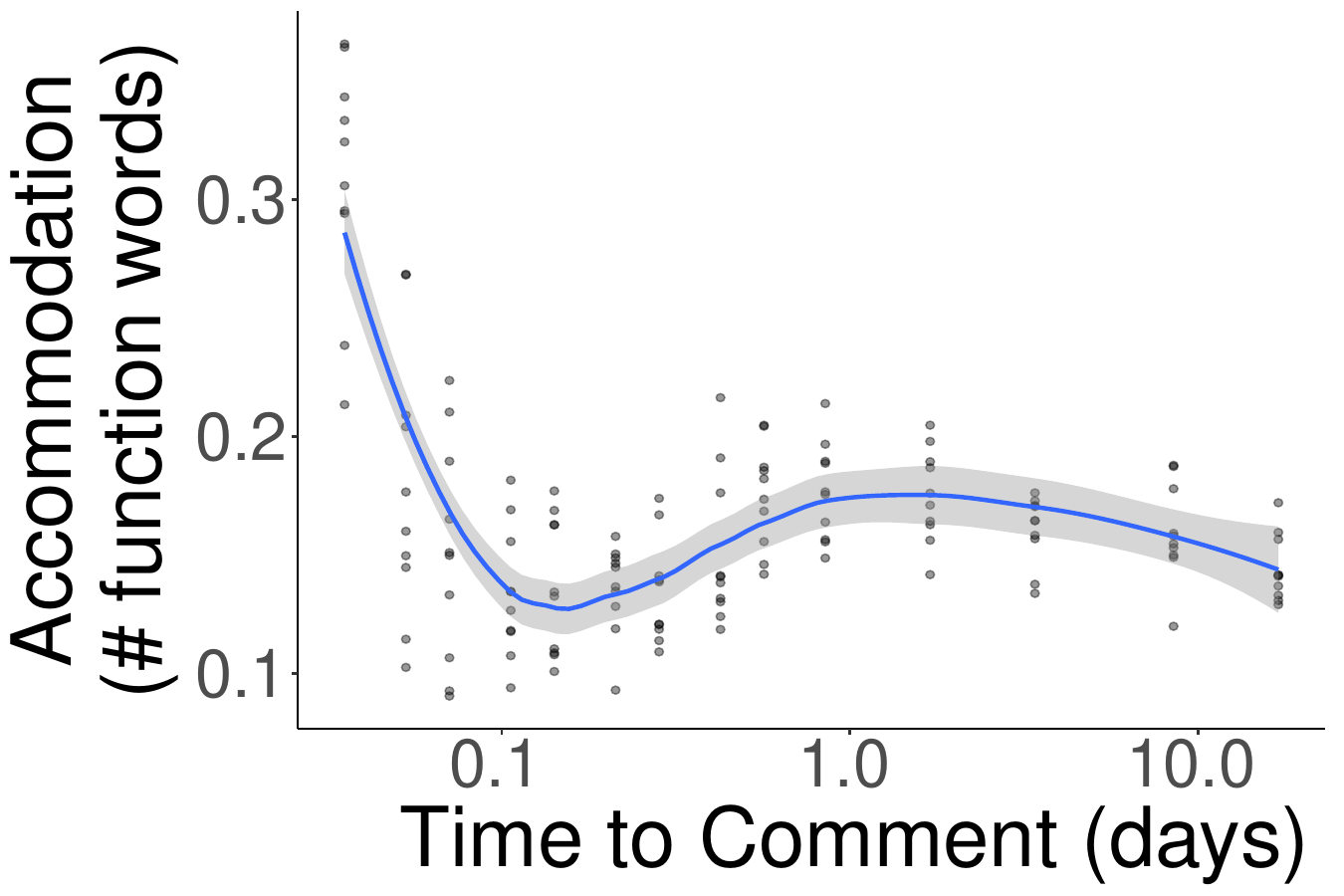}
        \caption{Time to Comment, \# Function Words}
        \label{function:time}
    \end{subfigure}    
    \hfill
    \begin{subfigure}[b]{0.23\textwidth}
        \centering
        \includegraphics[width=\textwidth]{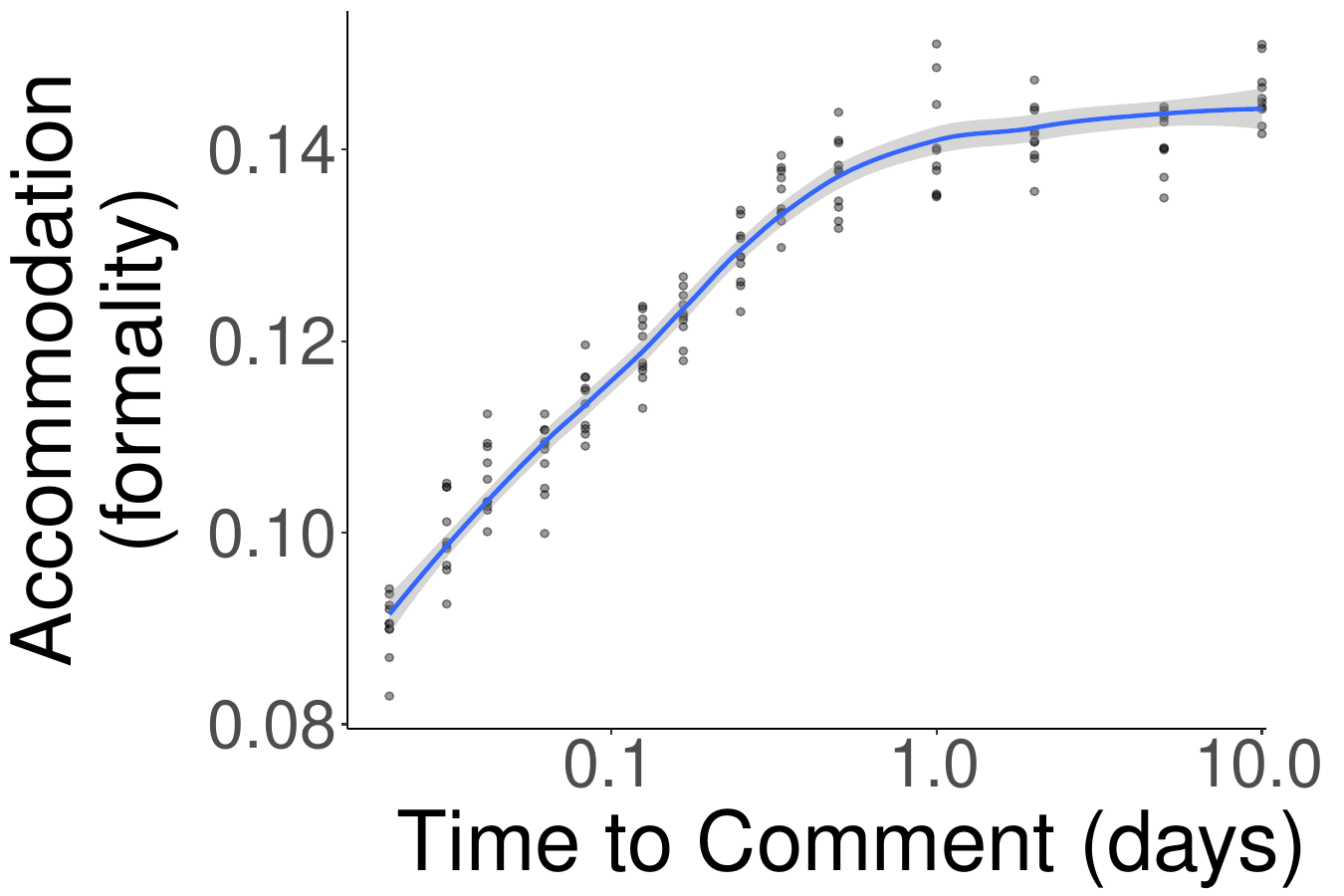}
        \caption{Time to Comment, Formality}
        \label{formality:time}
    \end{subfigure} 
    \caption{Commenters on Reddit are more likely to accommodate (a) \# function words when they reply quickly (suggesting subconscious accommodation) and b) formality of the comment when they reply slowly (suggesting strategic accommodation).}
    \label{fig:time-to-comment}
\end{figure}

Third, accommodation in the two style markers seems to occur via fundamentally distinct psychological processes. Accommodation can occur either 1) through a subconscious priming mechanism, where the speaker instinctively repeats what they hear; or 2) through a more conscious, strategic act with communicative intent \citep{doyle2016investigating}.
Figure~\ref{fig:time-to-comment} suggests that function-word-accommodation seems to be an unconscious form of relating to the audience, while formality-accommodation seems to be more intentional and strategic. Commenters exhibit greater accommodation in function words when they take less time to reply to the prior comment (\ref{function:time}) and greater accommodation in formality when they reply more slowly (\ref{formality:time}). These results are consistent with prior work, suggesting that accommodation of function words occurs subconsciously (reflexively, takes less time) and builds on this work to show that accommodation in other style markers, like formality, occurs strategically (intentionally, takes more time).

\begin{figure}[t!]
\centering
    \includegraphics[width=0.4\textwidth]{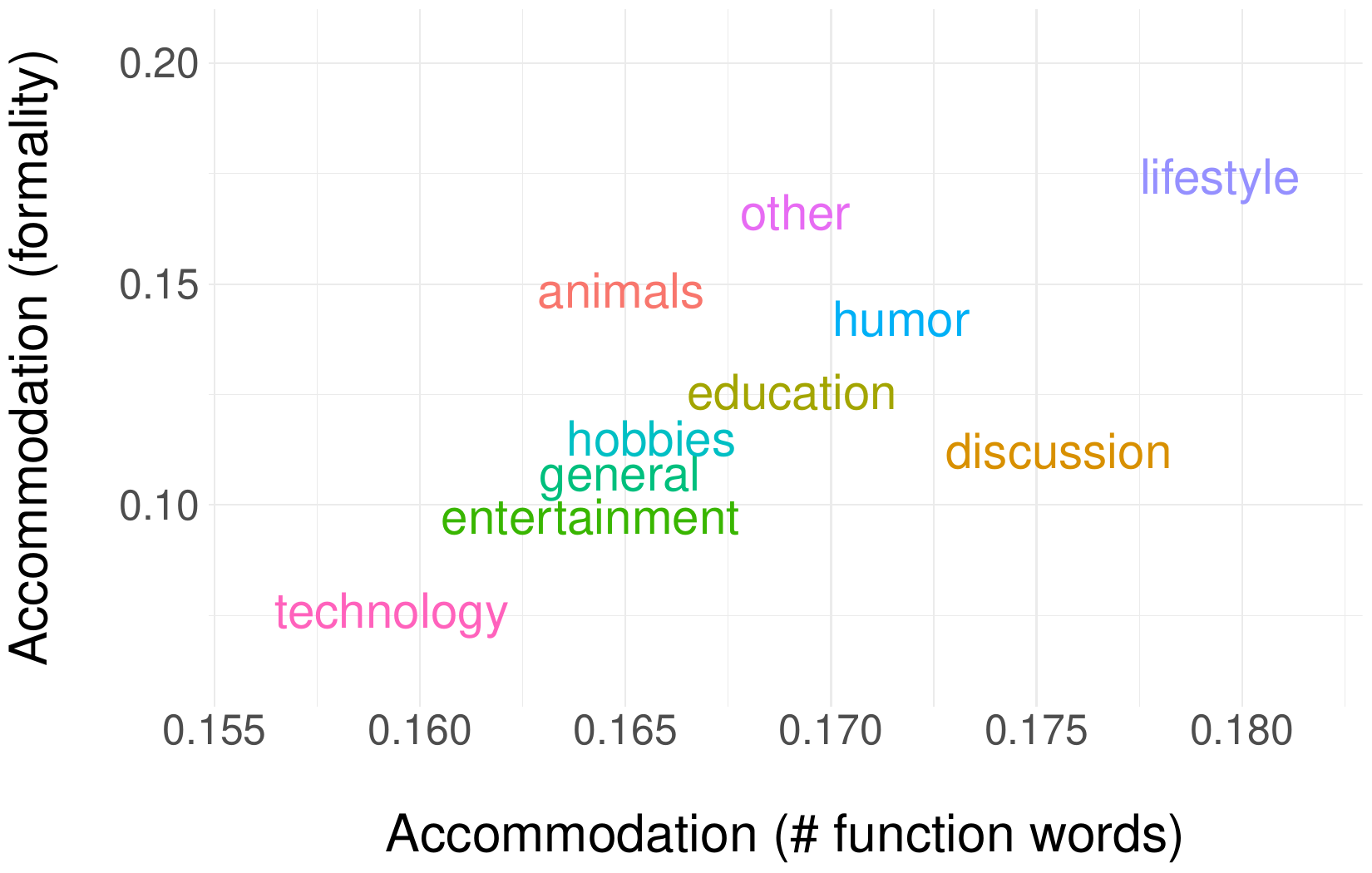}
    \caption{Level of accommodation in the number of function words (x-axis) and in formality (y-axis).}
    \label{fig:subreddit}
\end{figure}

\begin{figure}[t!]
\centering
    \includegraphics[width=0.4\textwidth]{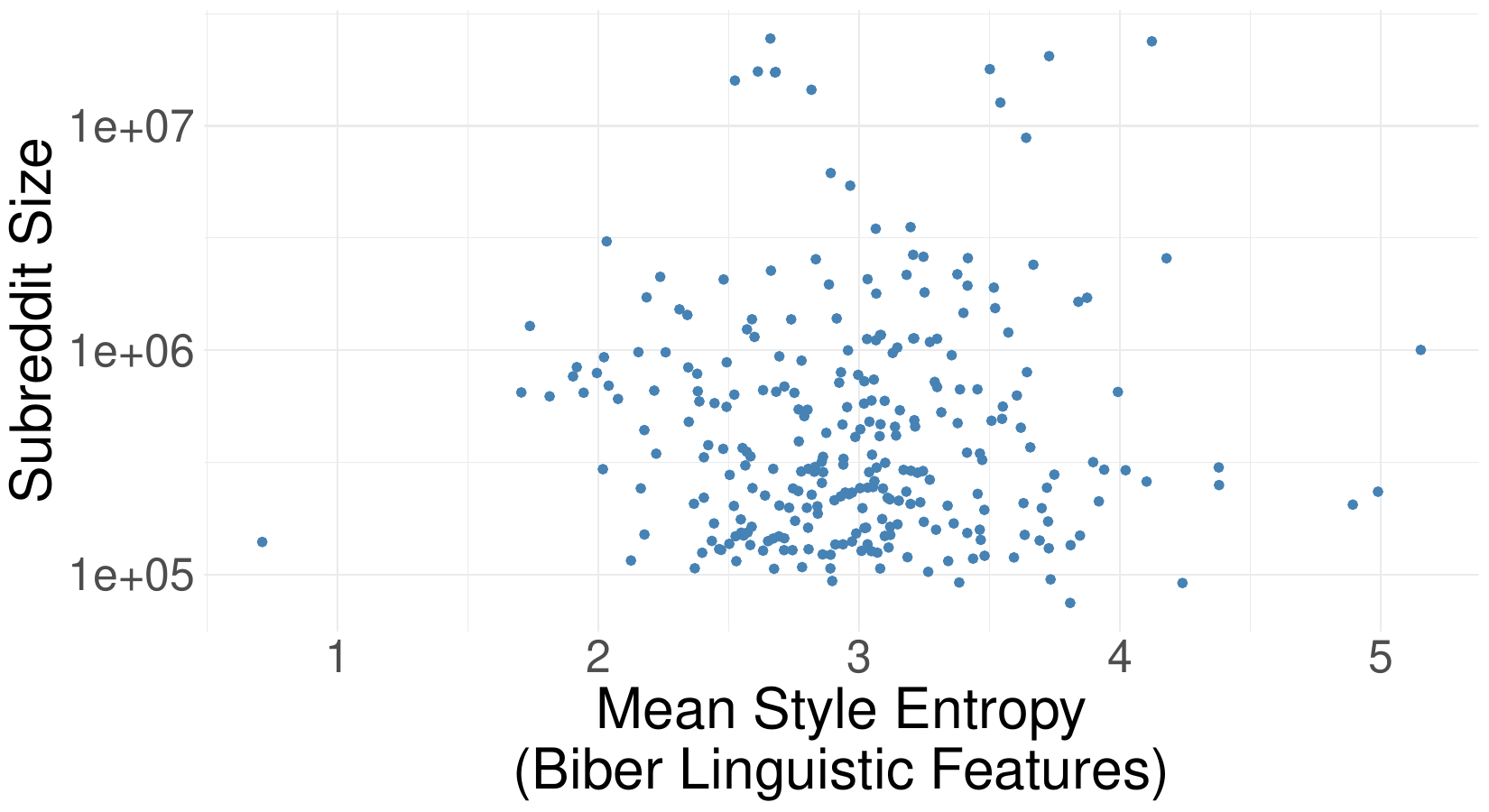}
    \caption{The mean Shannon Entropy of Biber's linguistic features (x-axis) is uncorrelated to the subreddit's number of subscribers (y-axis) ($p = 0.41$). Entropy is calculated using a random sample of comments in each subreddit.}
    \label{fig:entropy}
\end{figure}

Fourth, there is little variation in accommodation across subreddit characteristics. Figure~\ref{fig:subreddit} shows the levels of accommodation across ten different types of subreddits, using an existing taxonomy of popular subreddits.\footnote{\url{https://www.reddit.com/r/ListOfSubreddits/wiki/listofsubreddits/}} While certain types of subreddits (e.g., lifestyle) tend to have higher levels of accommodation than others (e.g., technology, entertainment), most differences are only weakly significant ($p>0.01$) with a small effect size. Moreover, Figure~\ref{fig:entropy} shows the relationship between subreddit size and variation in linguistic style, for 300 subreddits sampled based on their number of subscribers. To calculate variation in linguistic style, we use \citet{biber_1988}'s comprehensive set of linguistic features. 
Linguistic variation within each subreddit is estimated as the mean Shannon Entropy of each Biber tag frequency at the subreddit level. 
Despite expectations that larger communities may exhibit greater diversity in language use \citep{kocab2019size}, we find no relationship between community size and linguistic variation. 

Overall, these findings point to the nuanced dynamics of LSM in online interactions, indicating that factors such as function word usage and formality in the parent comment are associated with the linguistic style and tone of replies.

\section{What factors about a comment influence the degree of accommodation?}
LSM can be affected by many factors and existing studies have pointed out the roles of not only linguistic characteristics but also the contextual factors affecting LSM \citep{niederhoffer2002linguistic}. In this section, we study the connection between LSM and a series of contextual factors where the comment is posted (i.e., comment depth) and the ``success'' of a comment (i.e., comment Karma and parent comment Karma).


\begin{figure}[t]
\centering
    \begin{subfigure}[b]{0.23\textwidth}
        \centering
        \includegraphics[width=\textwidth]{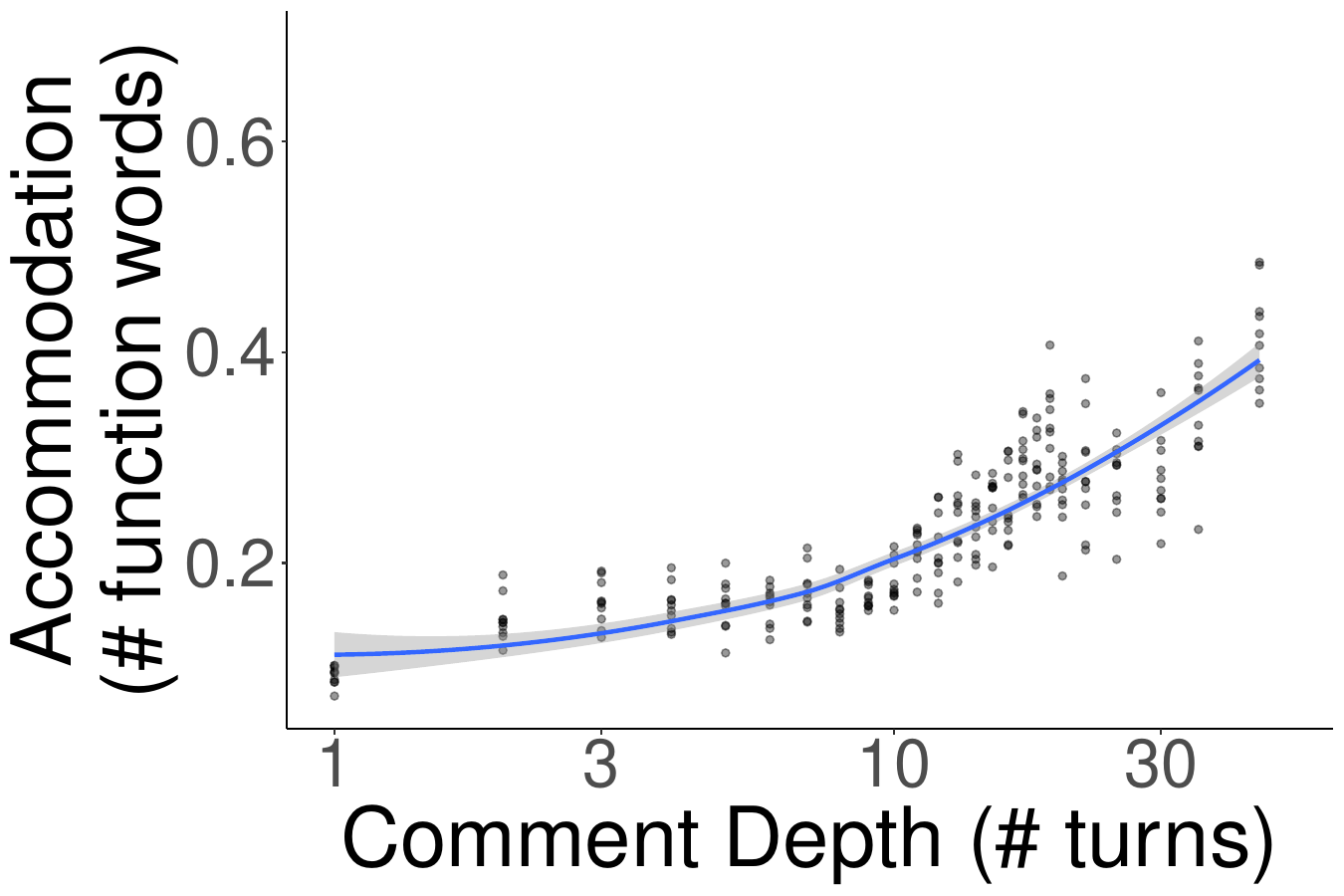}
        \caption{Comment Depth, Function Words}
        \label{function:depth}
    \end{subfigure}
    \hfill
    \begin{subfigure}[b]{0.23\textwidth}
        \centering
        \includegraphics[width=\textwidth]{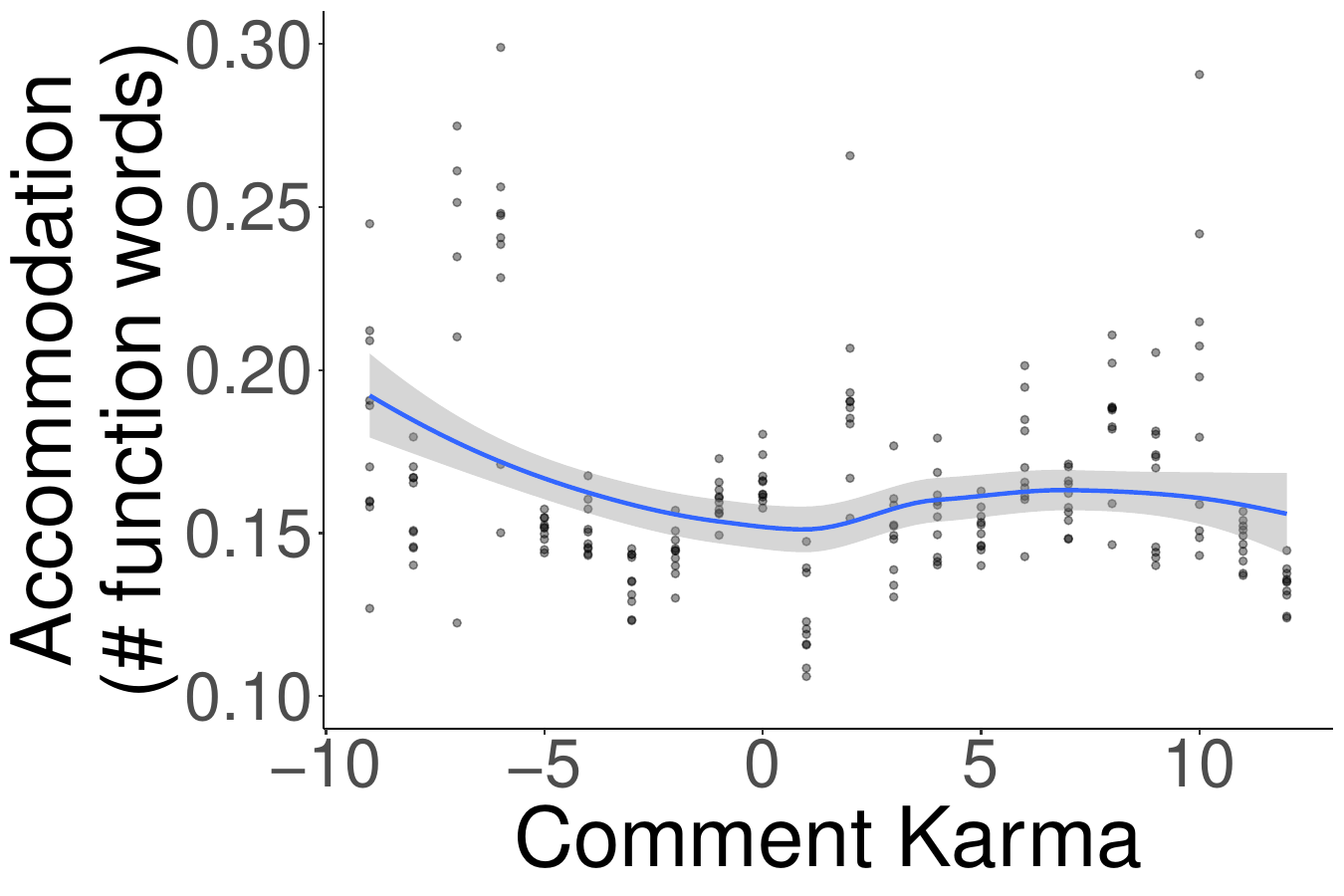}
        \caption{Comment Karma, Function Words}
        \label{function:karma}
    \end{subfigure}
    \hfill
    \begin{subfigure}[b]{0.23\textwidth}
        \centering
        \includegraphics[width=\textwidth]{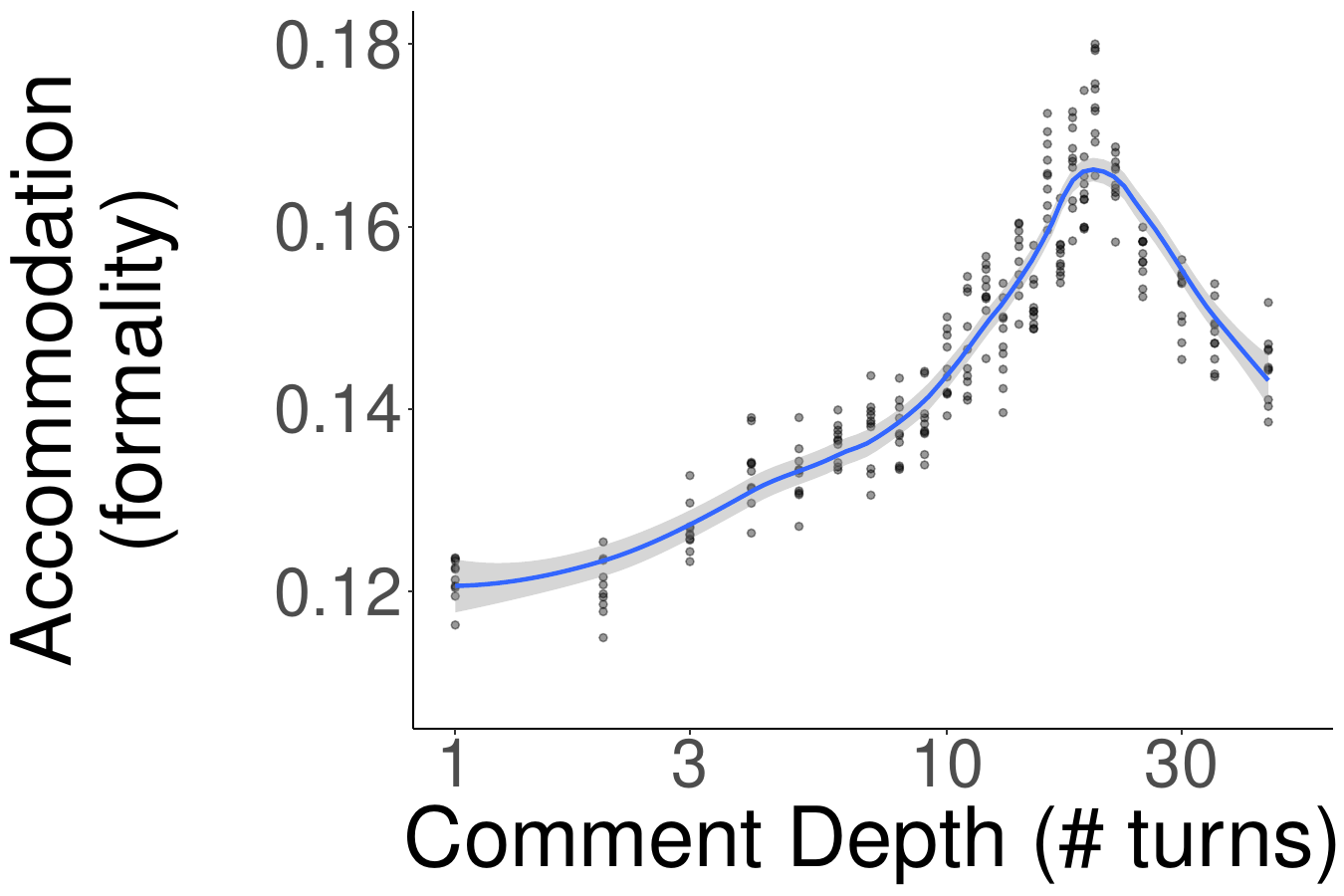}
        \caption{Comment Depth, Formality}
        \label{formality:depth}
    \end{subfigure}
    \hfill
    \begin{subfigure}[b]{0.23\textwidth}
        \centering
        \includegraphics[width=\textwidth]{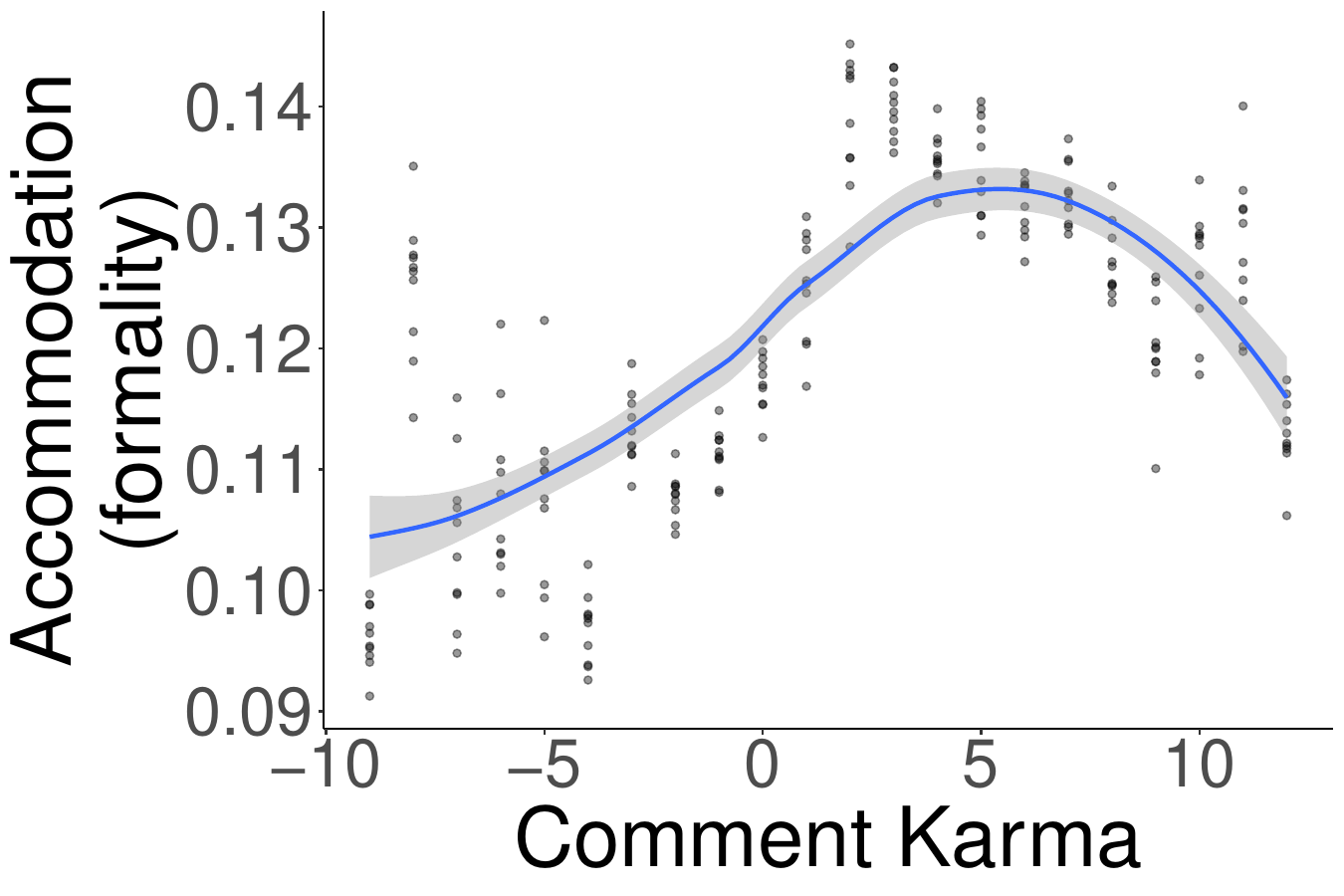}
        \caption{Comment Karma, Formality}
        \label{formality:karma}
    \end{subfigure}
    \caption{Characteristics of the comment are non-linearly associated with accommodation, including comment depth (a,c) and comment karma (b,d).}
    \label{fig:characteristics}
\end{figure}

\subsection{Experimental Setup}

To test for heterogeneity in the level of accommodation with respect to several covariates (e.g., depth, Karma), we run a mixed effects regression similar to Section~\ref{sec:measure}.2, but include an interaction term to test whether accommodation changes significantly with respect to some covariate (say, Karma $K$):

\begin{align*}
\label{reg_accommodation_interact}
m_b \sim & \beta_0 + \beta_1 m_a + \beta_2 K + \beta_3 m_a*K + \\
& \beta_4 L_a + \beta_5 L_b + \beta_6 r_{b \rightarrow a} + (1 | b) + (1 | s)
\end{align*}

Here, $\beta_1$ measures the level of accommodation when $K=0$ and $\beta_3$ measures the increase in accommodation when $K$ increases by one point; if $\beta_3$ is significantly different from $0$, then we have evidence that accommodation is heterogeneous with respect to Karma. 

In order to visualize these effects, we fit the model in the above equation to estimate accommodation at different values of Karma. In order to appropriately represent uncertainty in this model, we sample 100,000 conversation turns at each value of Karma 10 times and use this to obtain 10 different estimates of accommodation for each value of the covariate. To visualize the association between Karma and accommodation, we plot Karma on the x-axis and the LSM estimates on the y-axis.

\subsection{Results}

As shown in Figure~\ref{fig:characteristics}, various factors of comments are related to LSM.


\paragraph{Comment depth}
Comment depth reflects the position of a comment in the conversation tree. Deeper comments are usually posted in longer conversations and when the users are more engaged in the dialogue. As shown in Figure~\ref{function:depth} and Figure~\ref{formality:depth}, comment depth is positively correlated with LSM. However, accommodation in formality drops off for very deep comments. LSM happen more when the comment is deeper in the conversation tree, suggesting that users tend to match not only the content but also the structural aspects of their language in response to their interlocutor. Such a trend could be due to greater investment in the conversation. When two users are involved in longer and deeper conversations, they are more likely to be engaged in the conversation, which may lead to higher subconscious but lower conscious LSM. 


\paragraph{Comment Karma}

A key feature of Reddit is the ability for users to upvote or downvote comments, which determines the comment's karma - a measure of its popularity within the community. In figure \ref{fig:characteristics}, we observe several non-linear associations between karma, comment characteristics, and LSM. In terms of comment karma, users' LSM tends to remain relatively constant, except for cases where the comment has very high karma, which is associated with an increase in LSM. This finding implies that highly popular comments may foster greater linguistic alignment between users. 

We also see that comments with low karma have lower levels of LSM than comments with high karma (Figure~\ref{formality:karma}), which makes sense since we'd expect users to respond better to comments whether the author is mirroring their interlocutor. Notably, this upward trend reverses in comments with very high karma -- which have lower levels of LSM than comments with lower levels of karma. The reversal of the LSM trend in comments with high karma warrants further exploration. One possible explanation for this phenomenon is that highly upvoted comments may exhibit unconventional linguistic styles that deviate from the norm, which could be seen as novel by the Reddit community. Another explanation may be that comments with high karma are more likely to be popular in larger, diverse communities where users may have a wider range of linguistic styles. Additionally, it is possible that comments with high karma receive a higher volume of comments and interactions, which may dilute the overall LSM score due to the presence of diverse linguistic styles from multiple interlocutors.


\section{What effect does accommodation have on the conversation itself?}

Linguistic accommodation is usually associated with positive social benefits \cite{taylor2008linguistic, gonzales2010language}. Here, we test whether linguistic accommodation is associated with two positive behaviors in social media: sustained conversation and length of participation in a subreddit.


\subsection{Experimental Setup}

We fit a linear regression on conversational dyads following the LSM measure in Section~\ref{sec:measure}.2. Following the procedure from the prior section, we estimate the level of accommodation for comments around a particular covariate by sampling 100,000 conversation turns at or near the respective value of the covariate. Once again, we verify that differences between covariates are significant, by introducing interaction terms in the regression and testing for a statistically significance effect.

\subsection{Results}

\begin{figure}[t]
    \centering
    \begin{subfigure}[b]{0.23\textwidth}
        \centering
        \includegraphics[width=\textwidth]{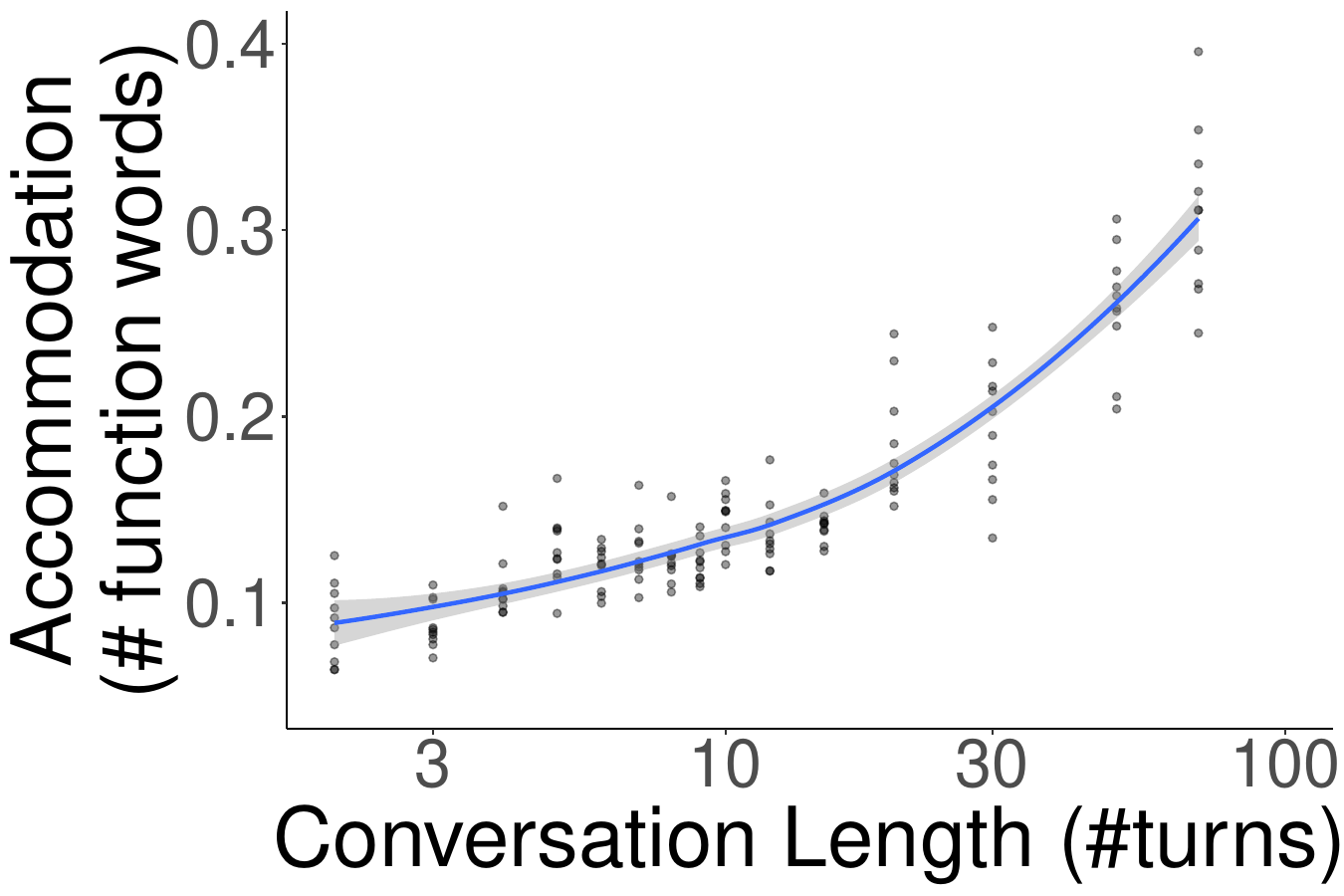}
        \caption{Length, \# Function Words}
        \label{length:functionwords}
    \end{subfigure}
    \hfill
    \begin{subfigure}[b]{0.23\textwidth}
        \centering
        \includegraphics[width=\textwidth]{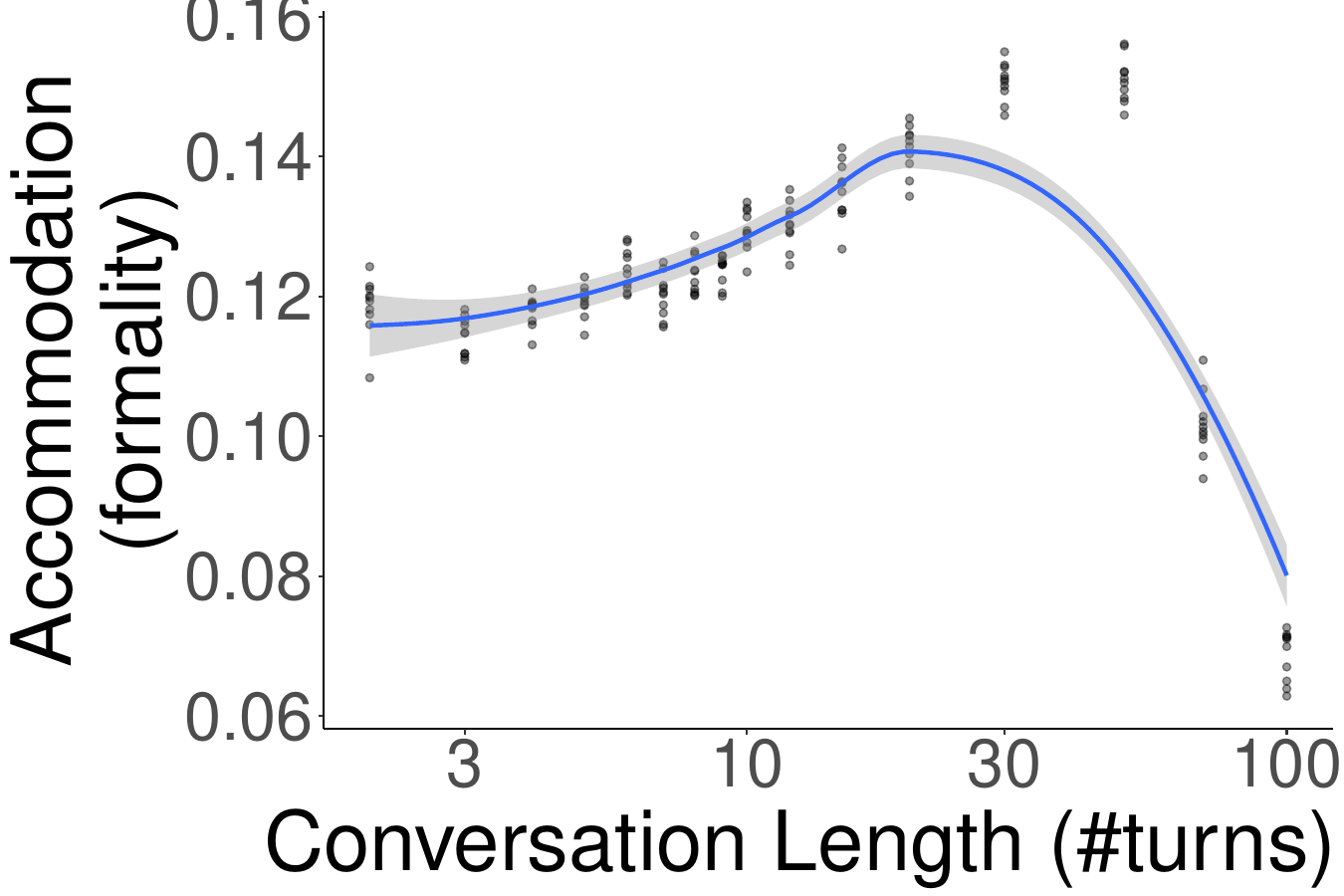}
        \caption{Length, Formality}
        \label{length:formality}
    \end{subfigure}
    \hfill
    \begin{subfigure}[b]{0.23\textwidth}
        \centering
        \includegraphics[width=\textwidth]{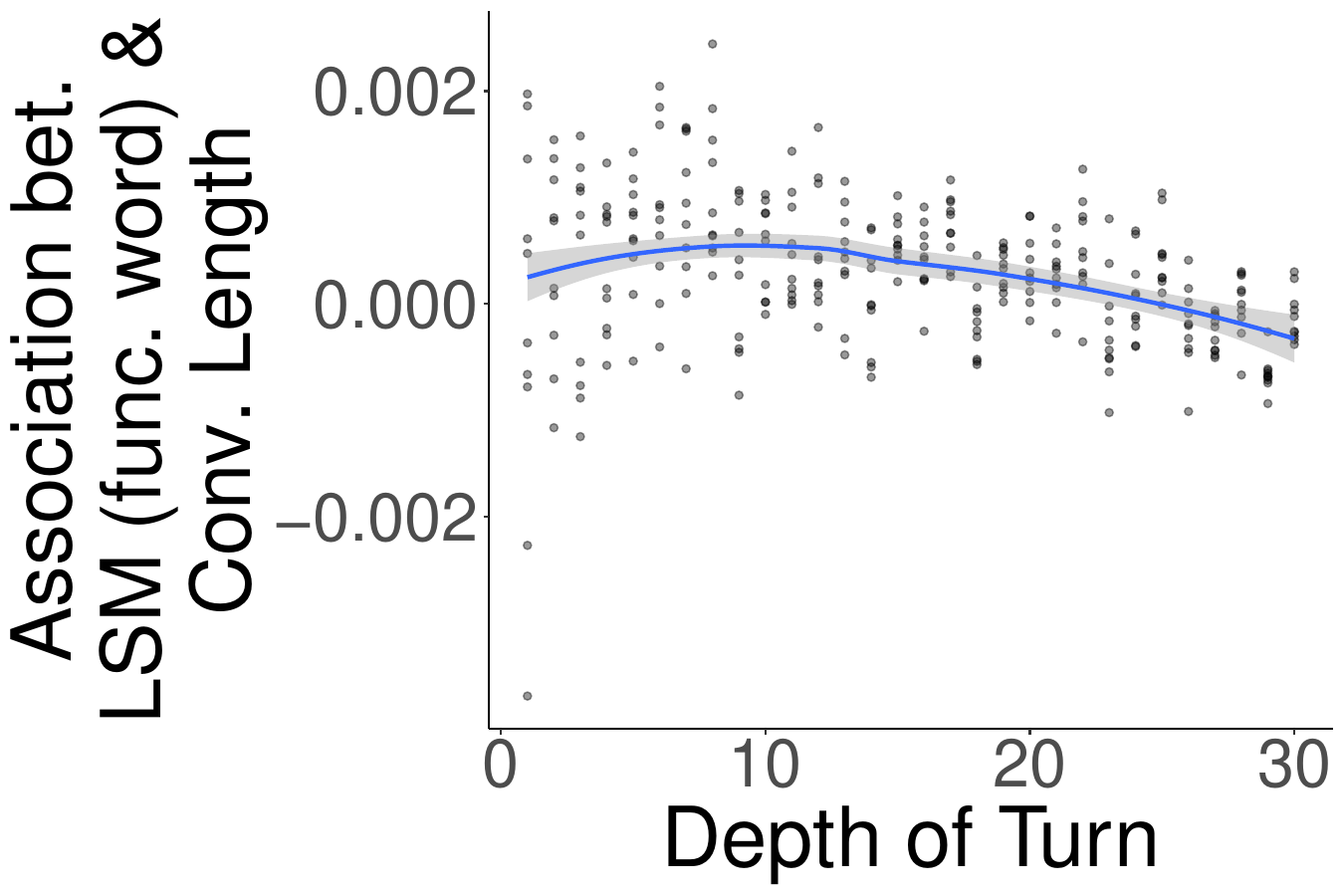}
        \caption{Depth x Length, \# Function Words}
        \label{depthlength:functionwords}
    \end{subfigure}
    \hfill
    \begin{subfigure}[b]{0.23\textwidth}
        \centering
        \includegraphics[width=\textwidth]{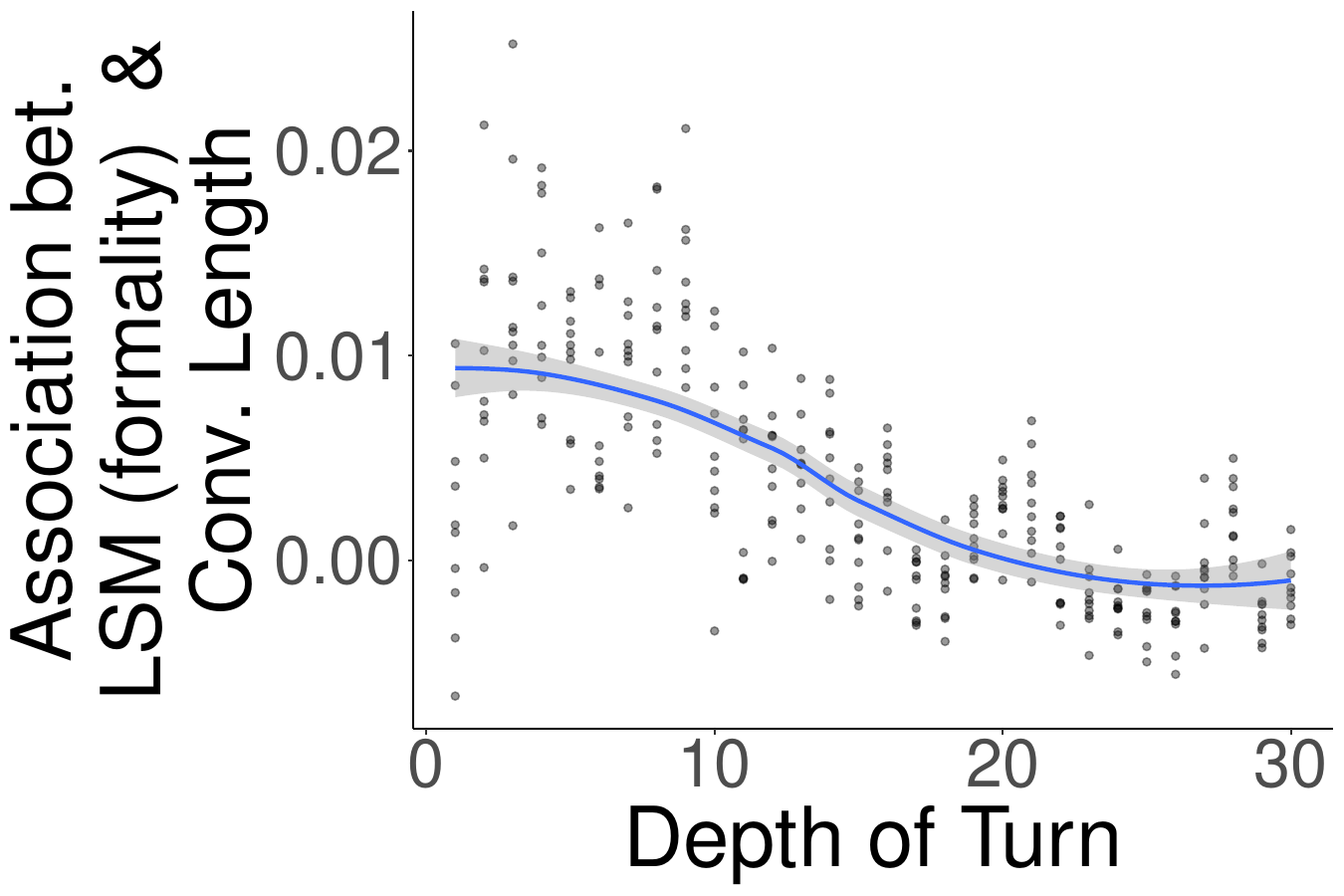}
        \caption{Depth x Length, Formality}
        \label{depthlength:formality}
    \end{subfigure}
    \caption{The number of turns in a conversation is associated with the level of accommodation in each turn: (a-b) Longer conversations (i.e., threads consisting of more conversation turns) are associated with higher accommodation, up to a threshold; for sufficiently long conversations, the association is negative for formality. (c-d) The effect of alignment on conversation length is stronger earlier in the conversation and weaker as more conversational turns occur.}
    \label{fig:max_depth}
\end{figure}

Figures~\ref{fig:max_depth}a and \ref{fig:max_depth}b compare the effect of alignment when conditioned on the total length of the conversation thread. For both functions words and formality, we observe from the fitted lines that accommodation is more likely to happen from longer conversations, but only up to a certain length of approximately 30-40. This suggests the possibility of LSM being an earlier indicator of how engaged the users will be in a conversation. On the other hand, the likelihood of accommodation in formality decreases when the conversation becomes longer than a certain threshold, which suggests that speakers may stop consciously trying to accommodate once the conversation becomes sufficiently long. 

Figures~\ref{fig:max_depth}c and \ref{fig:max_depth}d compare accommodation likelihoods at a given turn within a conversation. Interestingly, we can observe that LSM starts off highest at the beginning of a conversation and decreases as the number of turns increases. Combining the two results, we can conjecture that while the degree of LSM generally decreases within a conversation thread, the initial levels of LSM observed at the early stages of a conversation can indicate how engaged the speakers will be, which one can use to estimate the overall conversation length.





\begin{figure}
    \centering
    \begin{subfigure}[b]{0.23\textwidth}
        \centering
        \includegraphics[width=\textwidth]{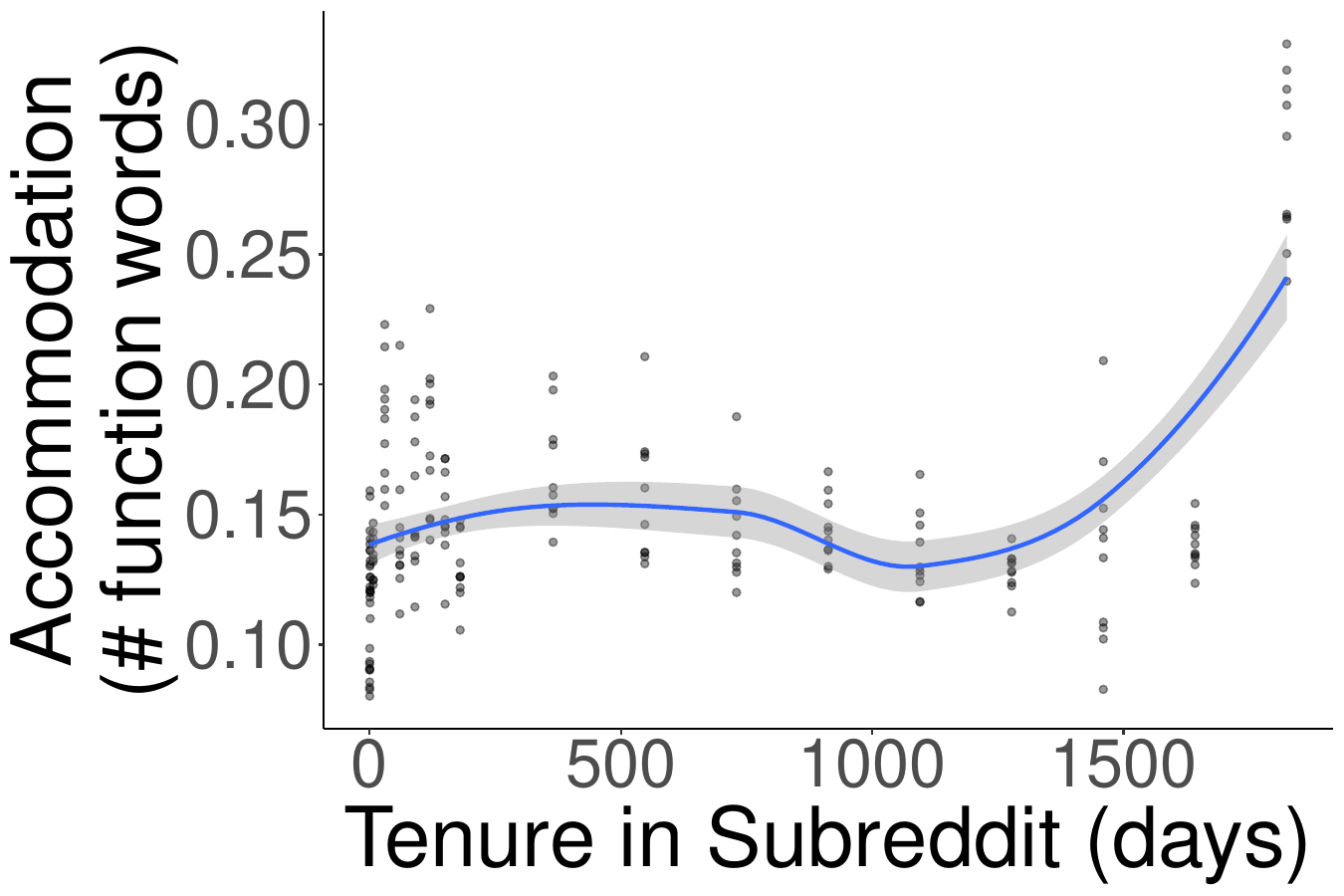}
        \caption{Tenure, \# Function Words}
        \label{tenure:functionwords}
    \end{subfigure}
    \hfill
    \begin{subfigure}[b]{0.23\textwidth}
        \centering
        \includegraphics[width=\textwidth]{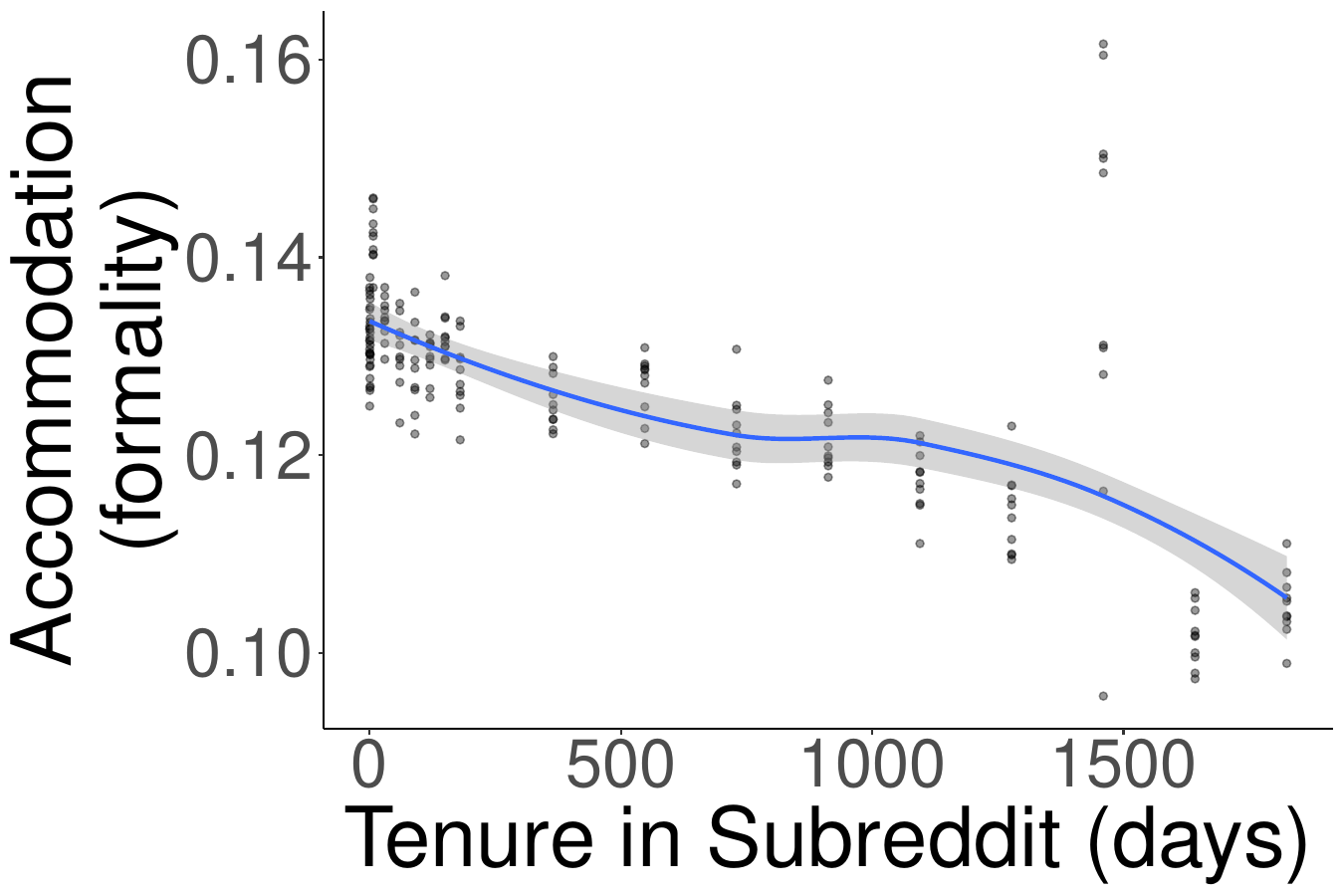}
        \caption{Tenure, Formality}
        \label{tenure:formality}
    \end{subfigure}    
    \caption{Alignment is associated with engagement: (a) stronger function-word LSM and (b) weaker formality LSM for higher tenure in subreddit.}
    \label{fig:tenure}
\end{figure}

How does LSM differ by tenure and number of subsequent posts in a subreddit? Figure~\ref{fig:tenure} shows that, for both style markers, users who have a longer tenure in the subreddit or who post more in the subreddit in the next month tend to display higher subconscious and lower conscious LSM. We consider these results as evidence of the ``lifespan'' of a user's engagement toward conversations held within that subreddit, and ultimately engagement toward the subreddit itself, which has been noted in prior work~\citep{danescu2013communities}.

\section{What effect does the social context have on accommodation: Controversiality?}

In this section, we examine whether LSM differs by social contexts that arise during conversations. Specifically, we focus on the controversy level of the parent comment.
In contrast to non-controversial issues, controversial issues lead to competitive disagreement, where the goal of the groups involved in argumentation is to convince the opponent group(s) of the validity of one’s point of view \cite{ilie2021discussion}. The arguments on controversial issues tend to invite strong emotions with negative affect \cite{mejova2014controversy} and deteriorate the deliberation in the public sphere because interactions often turn uncivil \cite{doxtader1991entwinement}. 


\subsection{Experimental Setup}

Following the procedure from the prior section, we estimate the level of accommodation for comments at each covariate, separately for controversial and non-controversial comments. 
When a comment or post receives a substantial number of upvotes and downvotes, Reddit automatically designates it as controversial. The exact method used by Reddit to determine controversy remains private. However, the Reddit API offers a binary label indicating whether a comment is controversial or non-controversial \cite{koncar2021analysis}. Approximately 1.30\% (n=218,899) of the comments in our sample are labeled as controversial.

We test that differences between conditions are significant with a three-way interaction term in the regression between the parent-comment style, the comment's Karma (or other covariates) and the comment's controversiality: $m_a \times K \times C$.






\subsection{Results}

\begin{figure}[t]
    \centering
    \begin{subfigure}[b]{0.22\textwidth}
        \centering
        \includegraphics[width=\textwidth]{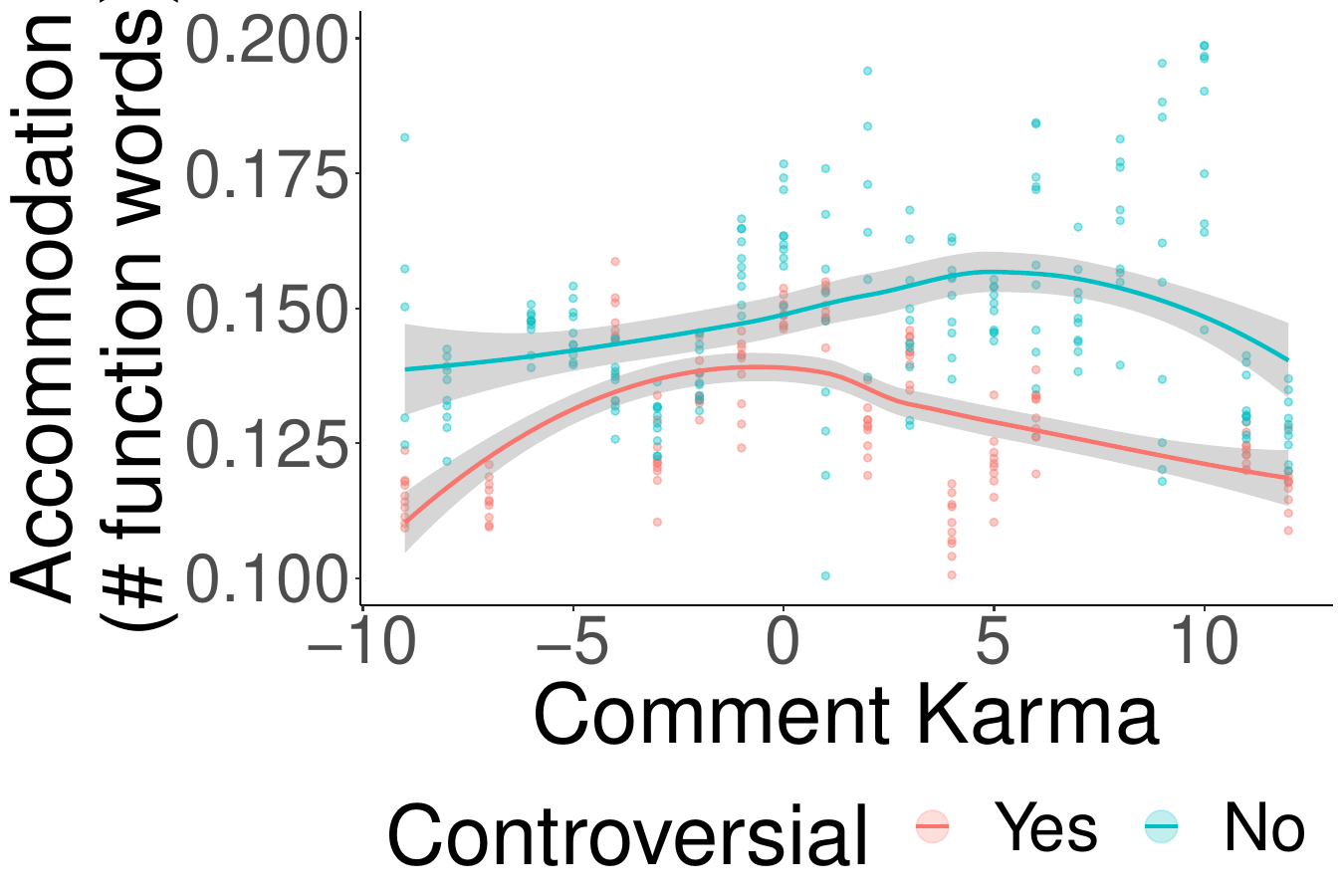}
        \caption{Karma, Function Words}
        \label{controversial:karma-func}
    \end{subfigure}
    \hfill
    \begin{subfigure}[b]{0.22\textwidth}
        \centering
        \includegraphics[width=\textwidth]{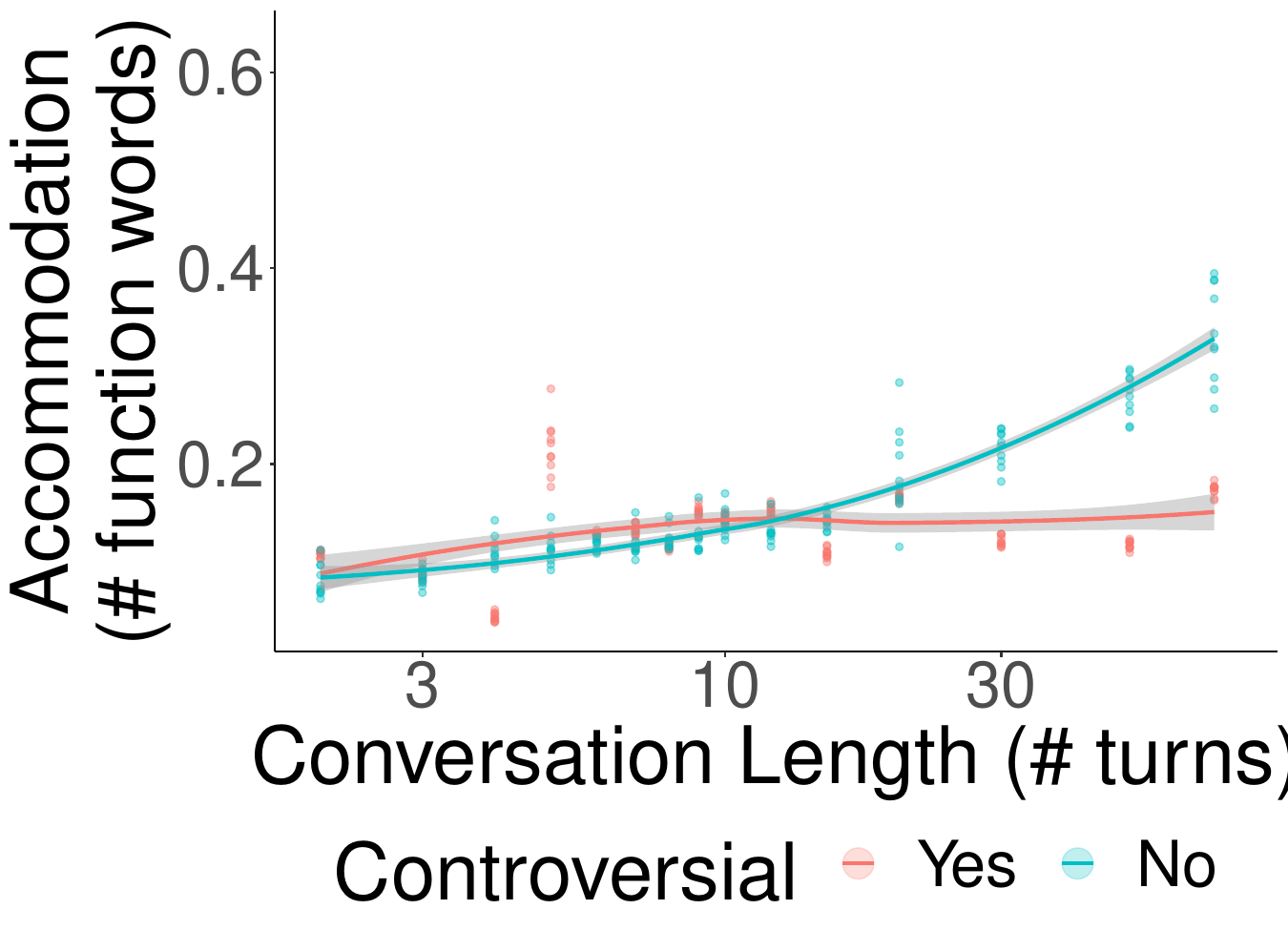}
        \caption{Length, Function Words}
        \label{controversial:length-func}
    \end{subfigure}
    \hfill
    \begin{subfigure}[b]{0.22\textwidth}
        \centering
        \includegraphics[width=\textwidth]{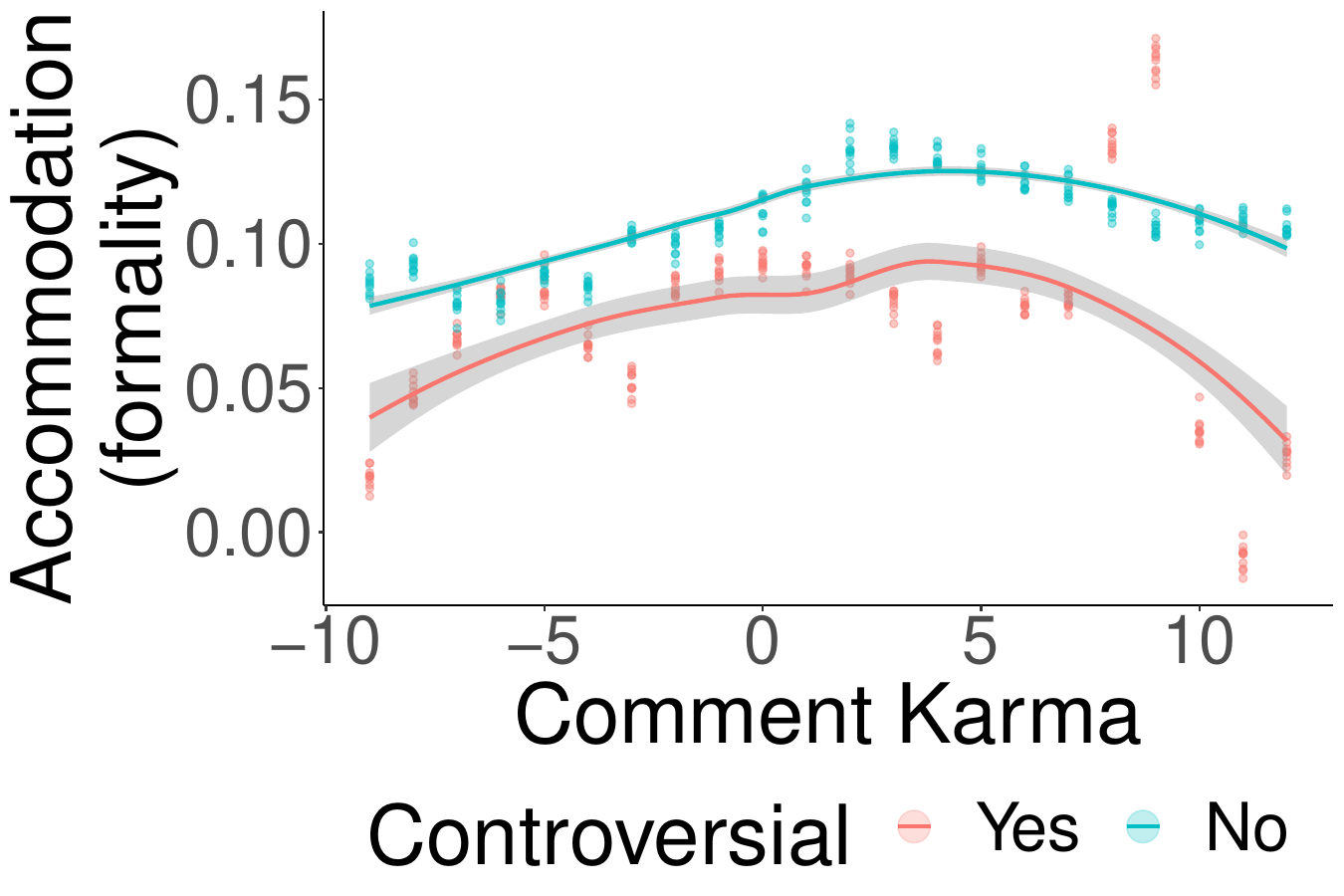}
        \caption{Karma, Formality}
        \label{controversial:karma}
    \end{subfigure}
    \hfill
    \begin{subfigure}[b]{0.22\textwidth}
        \centering
        \includegraphics[width=\textwidth]{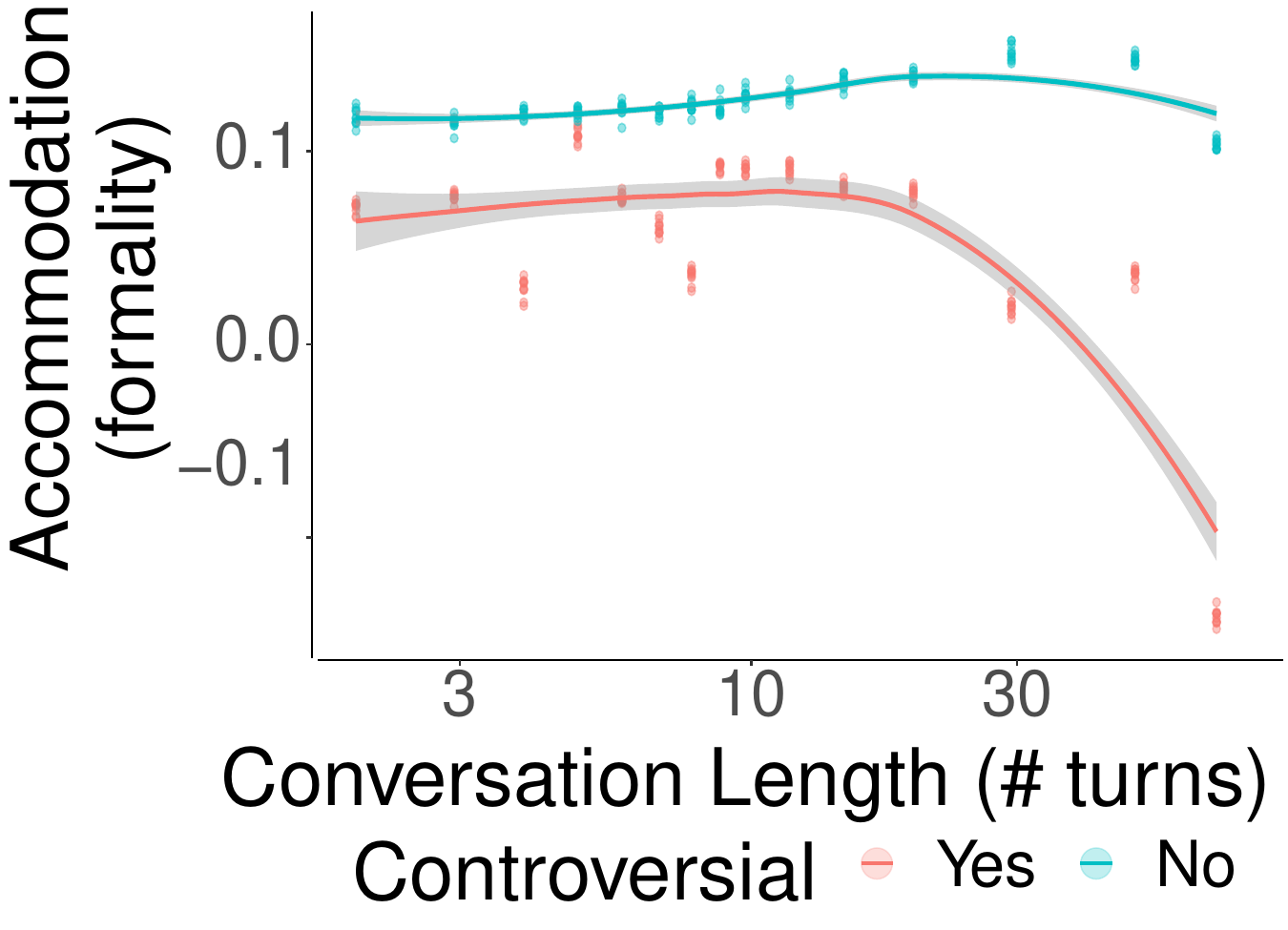}
        \caption{Length, Formality}
        \label{controversial:length}
    \end{subfigure}
    \caption{The associations between LSM and ``success'' are very different for controversial (red line) and non-controversial (blue line) comments: (a) Very-high-karma controversial comments tend to have lower, rather than higher, function-word alignment; and (c) overall lower formality-alignment. (b) Compared to shorter threads, longer controversial threads tend to have lower, rather than higher, function-word alignment and (d) formality alignment.}
    \label{fig:controversial}
\end{figure}

Figure~\ref{fig:controversial} reveals that LSM occurs differently in controversial and non-controversial comments. For both function words and formality, LSM is less likely to occur in controversial rather than non-controversial comments when the conversation length is below a certain threshold (12-14). Interestingly, we see that this trend is strengthened as the conversation length increases. One possible explanation is that controversial comments generate more initial interest that promotes users to engage more in conversations. However, this initial effect is washed away as the conversation takes further turns, and the conversation is less likely to continue due to reasons such as incivility. Non-controversial comments, on the other hand, enjoy less of this initial boost and is more likely to carry on if the users have accommodated each other's language during their conversation.

With the addition of Karma, we can observe a more complex trend that plays out differently for each style marker. For function words, conversations in controversial comments have a nonlinear relationship that drops as the parent comment's Karma increases, whereas a weak positive correlation can be observed for non-controversial comments and levels of Karma. In contrast, for formality, LSM occurs most at comments with about 0-5 Karma and decreases for higher Karma for both controversial and non-controversial comments.

Overall, we observe that social contexts that are defined by the community platform such as Karma or controversy have complex, nonlinear effects on how LSM occurs in conversations.






\section{Loss of Status via Community Banning}

Reddit bans specific subreddit communities as a result of policy violations, such as repeated posting of highly offensive content or lack of moderator oversight \cite{chandrasekharan2017you}. When users are highly active in such communities, the ban potentially results in a loss of status, as they are forced to find new communities to participate in. 
%
Here, we test the extent to which users change how they are linguistically influenced by others after such a ban. While prior work has studied how users change after \textit{gaining} status \citep[e.g.,][]{danescu2012echoes}, our unique setting allows us to perform a novel study of the potentially humbling effects of status loss. In addition,  a study of the subreddit \texttt{r/changemyview} suggests that formality is (weakly) associated with more effective persuasion on Reddit \citep{daytermesserli2022}; we hypothesize that users who recently experienced a ban may have multiple pragmatic reasons to accommodate more.

\subsection{Experimental Setup}

We test for changes to linguistic influence using a pseudo-causal difference-in-difference analysis \cite{lechner2011estimation}.  Subreddit ban dates were determined by identifying all banned subreddits and then using the last date of a post in that subreddit.  Our sample includes 1,024 subreddits banned between July 2019 and December 2022. We identify 16,686 users in our sample who made at least one comment in these subreddits in the 30 days before their ban. Each user from a banned subreddit is considered as treated and matched with a control user who did not participate in that subreddit. 

Three analyses of the effect of the ban are performed, controlling for user-level and temporal factors. First, we estimate the effect of commenting in a banned subreddit, by comparing posts made in banned subreddits $t$ months before the ban to posts made by the same users at the same time, in other subreddits. Second, using a difference-in-differences approach, we estimate the effect of banning a subreddit on authors' use of accommodation in (unbanned) subreddits they were active in for $t$ months before and after the ban. This second analysis measures the spill-over effects of the ban on users' behaviors in other subreddits; the difference-in-differences estimator uses users active in these subreddits at the same time, but not in a banned subreddit, as a control for temporal and subreddit-level effects. Third, we calculate the effect of the ban on commenting behavior in subreddits users migrated to (i.e., newly joined) after the ban was enacted. The difference-in-differences estimator compares accommodation in comments in the banned subreddits to comments in the subreddits these users migrated to; to isolate the effect of migration, the difference between the comments in the migrated and banned subreddits are compared against the spill-over effects in other subreddits that users were a part of during this time.

%

\subsection{Results}

\begin{figure}[t!]
    \centering
    \begin{subfigure}[b]{0.23\textwidth}
        \centering
        \includegraphics[width=\textwidth]{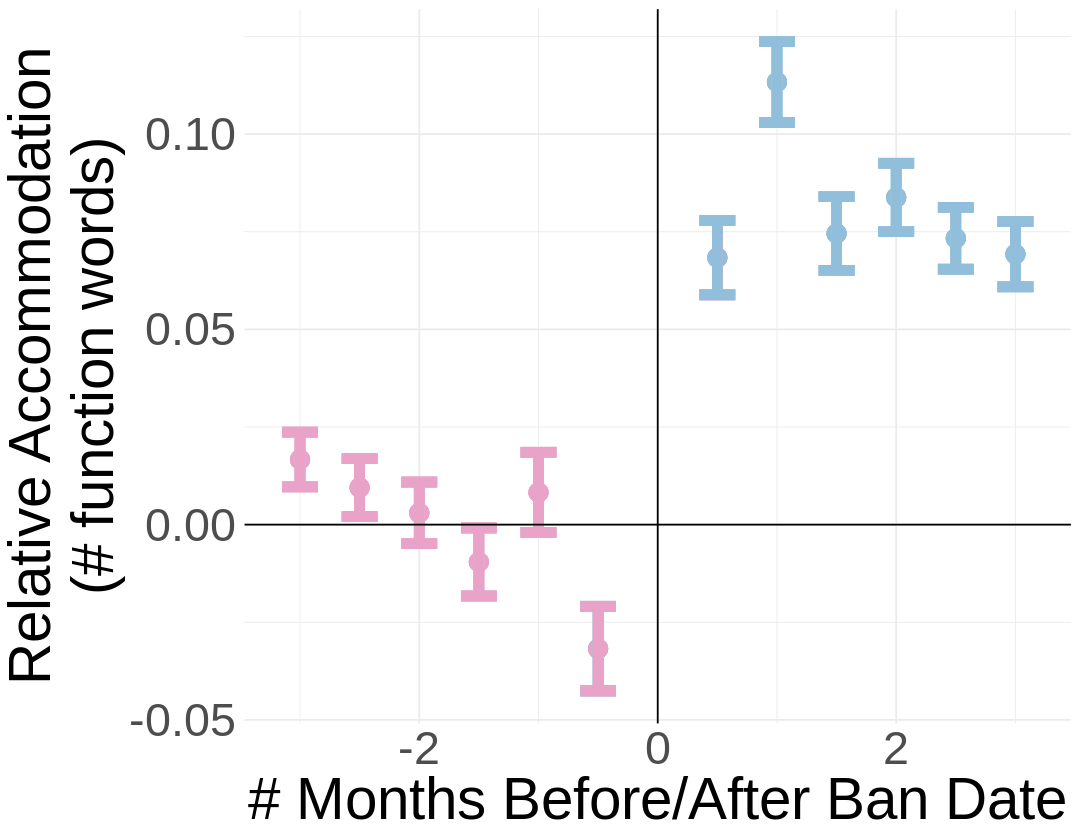}
        \caption{\# Function Words}
        \label{banned:function}
    \end{subfigure}
    \hfill
    \begin{subfigure}[b]{0.23\textwidth}
        \centering
        \includegraphics[width=\textwidth]{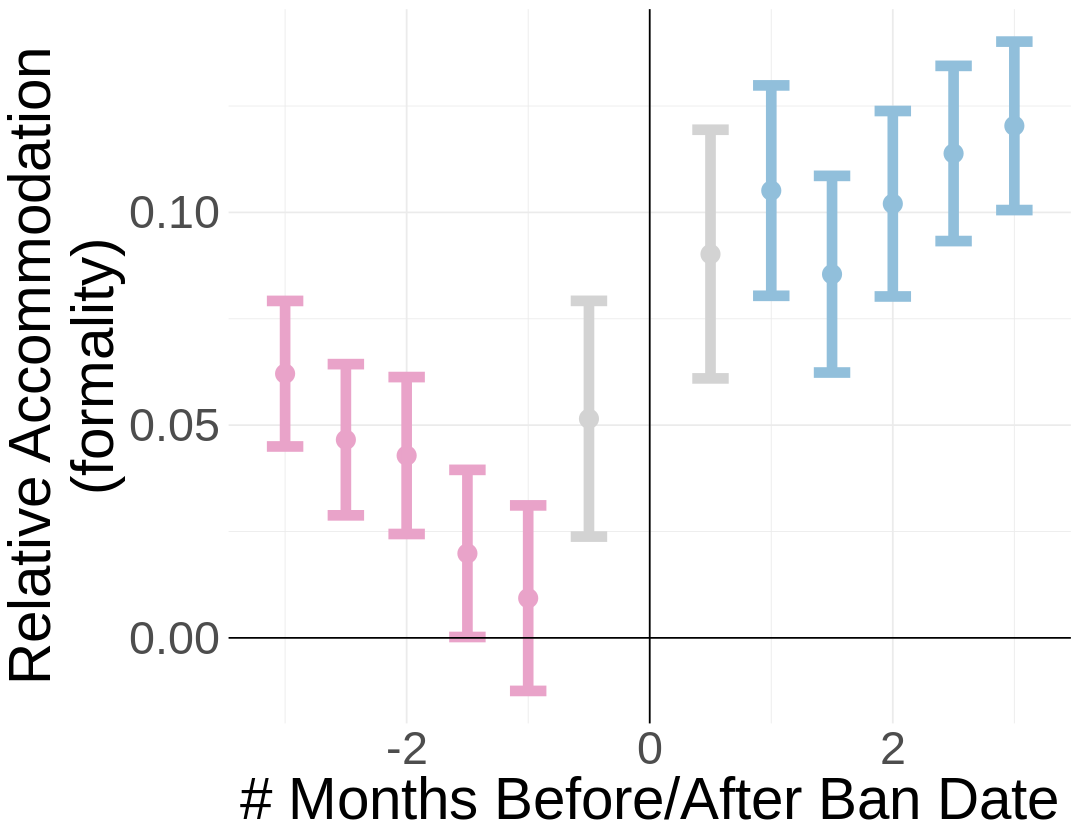}
        \caption{Formality}
        \label{banned:formality}
    \end{subfigure}  
    \hfill
        \begin{subfigure}[b]{0.23\textwidth}
        \centering
        \includegraphics[width=\textwidth]{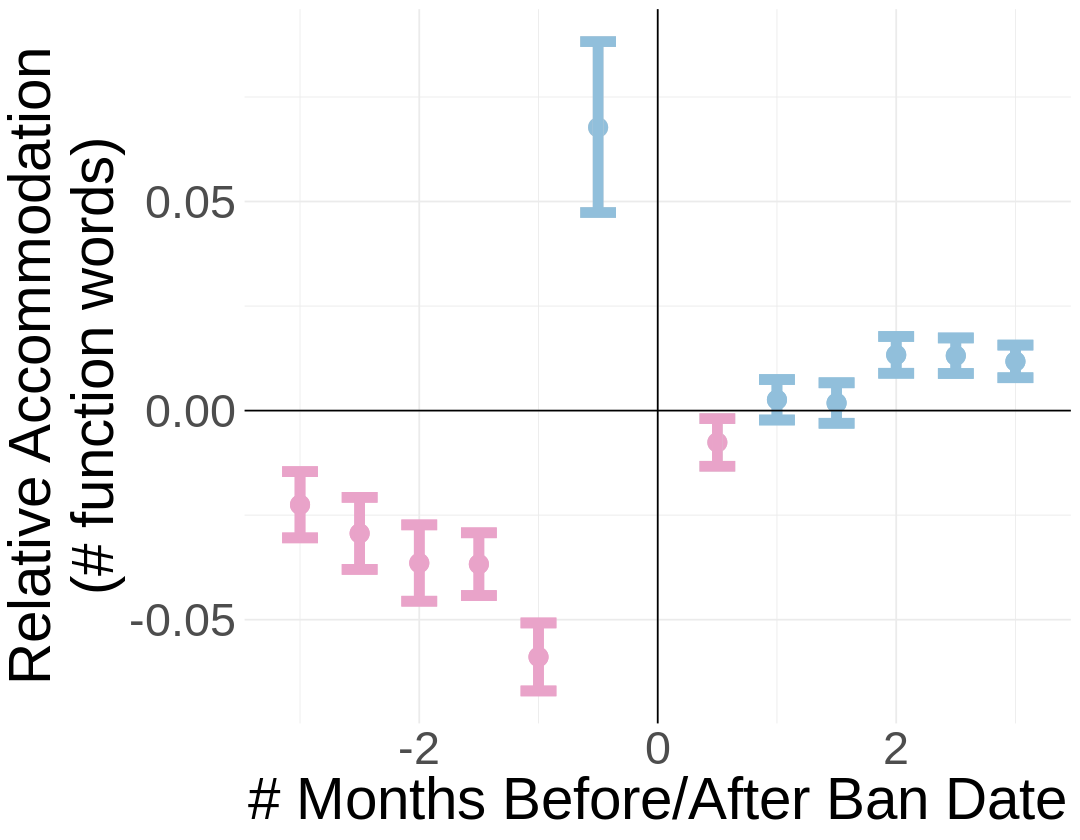}
        \caption{\# Function Words}
        \label{migrate:function}
    \end{subfigure}
    \hfill
    \begin{subfigure}[b]{0.23\textwidth}
        \centering
        \includegraphics[width=\textwidth]{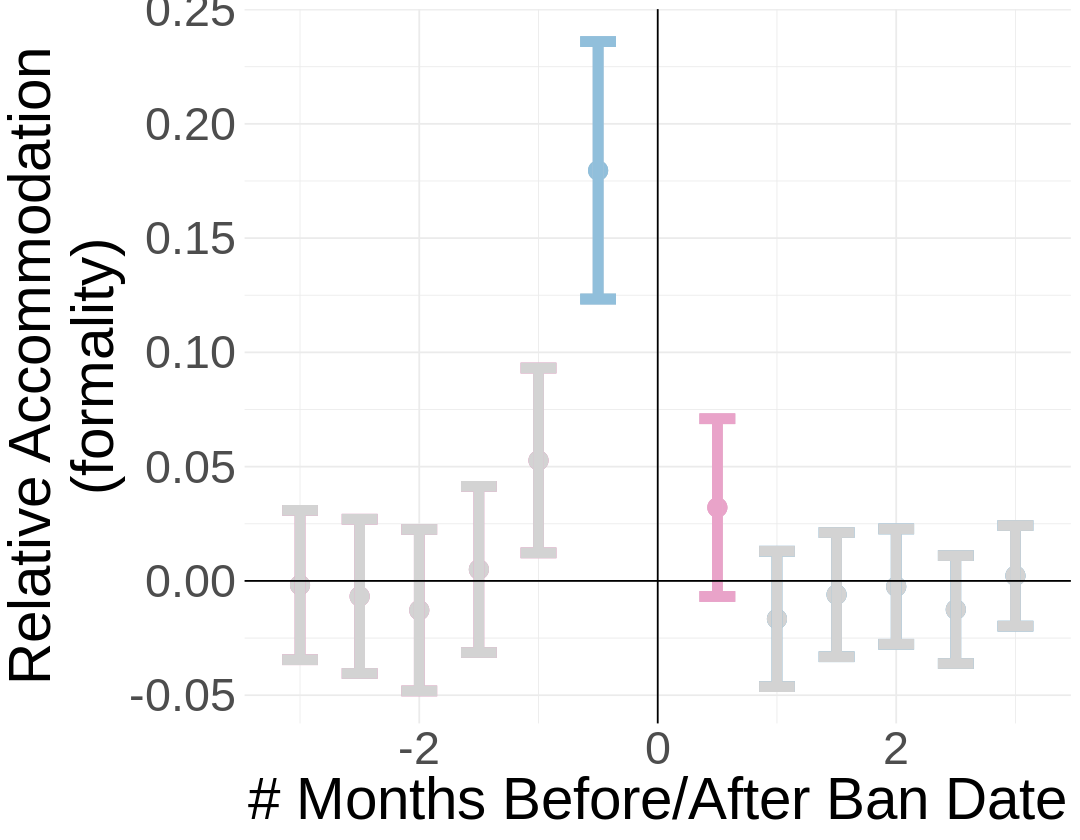}
        \caption{Formality}
        \label{migrate:formality}
    \end{subfigure}
    \caption{When a subreddit is banned, (a,c) users who commented there tend to have higher LSM in other subreddits, (b) users tend to have higher LSM in function words in subreddits they migrate to, (d) LSM in formality tends to temporarily increases just before the subreddit is banned then returns to near-baseline levels in subreddits they migrate to.}
    \label{fig:banned}
\end{figure}

Our results suggest that policy actions on Reddit, such as banning, have an effect on the level of accommodation by users. First, the level of subconscious accommodation tends to be lower in banned subreddits than other subreddits the users comment in during the 30 days before the ban (the effects are all below 0 in Figure ~\ref{migrate:function} ($p < 2e-16$). 

Second, following the banning of a subreddit, users tend to change their LSM levels in other subreddits: Figure~\ref{fig:banned} shows that function-word-mirroring (\fref{banned:function}) and formality-mirroring (\fref{banned:formality}) increase after a subreddit is banned. Our results suggest that users who had previously been active in banned subreddits may have been making an effort to index agreeableness by accommodating (e.g., to avoid losing status in another community). 

Third, changes in accommodation are initially amplified in subreddits that these users migrate to after their original community was banned. The comments left by these users in banned subreddits exhibit higher levels of accommodation than would be expected immediately before the ban and maintain higher subconscious accommodation in subreddits they migrated (Figures~\ref{migrate:function} and \ref{migrate:formality}$p < 2e-16$). Since function-word mirroring is likely subconscious and formality-mirroring strategic (Section~\ref{sec:measure}), our results suggest that users who had previously been active in banned subreddits may have, intrinsically, indexed agreeableness by accommodating (e.g., to gain status in their new community) but without making a conscious effort (e.g., because they were upset about the loss of a status). These users also increased LSM in the subreddit immediately before it was banned (e.g., perhaps to index agreeableness when warnings about the ban were issued).

\section{Discussion and Conclusion}
In this study, we performed a large-scale computational analysis on Reddit conversations to understand when LSM occurs and its effect on platform engagement. Overall, do our findings indicate that LSM frequently occurs in online conversations within Reddit, and that it exhibits complex nonlinear relationships with conversation metrics such as Karma, conversation lengths, or controversy scores, which suggests linguistic influence can affect conversation dynamics. Furthermore, we show that the degree of accommodation in conversations is related to greater levels of engagement both at conversation and platform levels. Our findings highlight the possibility of identifying LSM as an indicator of engagement and civil conversations and suggest ideas for building and maintaining online communities that promote constructive discourse. 

In our experiments, we have assumed LSM as a unidirectional concept by measuring the exhibition of a particular style conditioned on the previous turn. However, LSM can occur in several different directions, such as the two speakers converging into a single style or even diverging to separate styles. While not in the scope of this study, the existence of such types of LSM in Reddit conversation threads can be studies in future research.

\section{Ethical Considerations}
This study was conducted only on observational data and did not require any human intervention. We did not use any information that could identify individuals or specific demographic groups, and all of our presented results were obtained through aggregation from millions of users and comments.

\section*{Acknowledgments}

This material is based in part upon work supported by the National Science Foundation under Grant No IIS-2143529.

\bibliography{references.bib}
\bibliographystyle{acl_natbib}




\end{document}